\documentclass[11pt]{article}

\usepackage[final]{acl}

\usepackage{times}
\usepackage{latexsym}
\usepackage{enumitem}
\usepackage[T1]{fontenc}

\usepackage[utf8]{inputenc}

\usepackage{microtype}
\usepackage{hyperref}

\usepackage{inconsolata}
\usepackage{amsmath}
\usepackage{amsfonts}
\usepackage{graphicx} 
\usepackage{cleveref}
\usepackage{multirow}
\usepackage{multicol}
\usepackage{tcolorbox}
\usepackage[table]{xcolor}
\usepackage{xcolor}
\usepackage{booktabs}
\usepackage{todonotes}
\usepackage{stfloats}
\usepackage{subcaption}

\title{Controlling Distributional Bias in Multi-Round LLM Generation via KL-Optimized Fine-Tuning}

\author{
 \textbf{Yanbei Jiang\textsuperscript{1}},
 \textbf{Amr Keleg\textsuperscript{2}},
 \textbf{Ryandito Diandaru\textsuperscript{2}},
 \textbf{Jey Han Lau\textsuperscript{1}},
\\
 \textbf{Lea Frermann\textsuperscript{1}},
 \textbf{Biaoyan Fang\textsuperscript{3}},
 \textbf{Fajri Koto\textsuperscript{2}}
\\
\\
 \textsuperscript{1}University of Melbourne,
 \textsuperscript{2}Mohamed bin Zayed University of Artificial Intelligence,
 \textsuperscript{3}Oracle
\\
 \small{
   \textbf{Correspondence:} \href{mailto:yanbeij@student.unimelb.edu.au}{yanbeij@student.unimelb.edu.au}
 }
}

\begin{document}
\maketitle
\begin{abstract}
While the real world is inherently stochastic, Large Language Models (LLMs) are predominantly evaluated on single-round inference against fixed ground truths. In this work, we shift the lens to \textit{distribution alignment}: assessing whether LLMs, when prompted repeatedly, can generate outputs that adhere to a desired target distribution, e.g. reflecting real-world statistics or a uniform distribution. We formulate distribution alignment using the attributes of gender, race, and sentiment within occupational contexts. Our empirical analysis reveals that off-the-shelf LLMs and standard alignment techniques, including prompt engineering and Direct Preference Optimization, fail to reliably control output distributions. To bridge this gap, we propose a novel fine-tuning framework that couples Steering Token Calibration with Semantic Alignment. We introduce a hybrid objective function combining Kullback-Leibler divergence to anchor the probability mass of latent steering tokens and Kahneman-Tversky Optimization to bind these tokens to semantically consistent responses. Experiments across six diverse datasets demonstrate that our approach significantly outperforms baselines, achieving precise distributional control in  attribute generation tasks. Code and data are available at \url{https://github.com/YanbeiJiang/Distribution-Debias}.

\end{abstract}

\section{Introduction}

Although generative Large Language Models (LLMs) are inherently probabilistic, existing evaluation methods typically focus on a single-round generation \cite{de2019bias, soundararajan-delany-2024-investigating, pan-etal-2025-survey}. This narrow view becomes problematic in real-world applications, where repeated prompting can lead to distributional properties in LLM output that deviates from the intended statistics. We refer to this phenomenon as \textit{distribution bias} \cite{shrestha-srinivasan-2025-llm}, defined as the difference between the observed output distribution and the target distribution. Distribution bias matters because it influences how LLMs represent social attributes such as gender, race, and sentiment, with implications for fairness \cite{gallegos-etal-2024-bias}, personalization \cite{salemi-etal-2024-lamp}, and trustworthiness \cite{litschko-etal-2023-establishing}.

\begin{figure}[t] 
    \centering
    \includegraphics[width=0.85\linewidth]{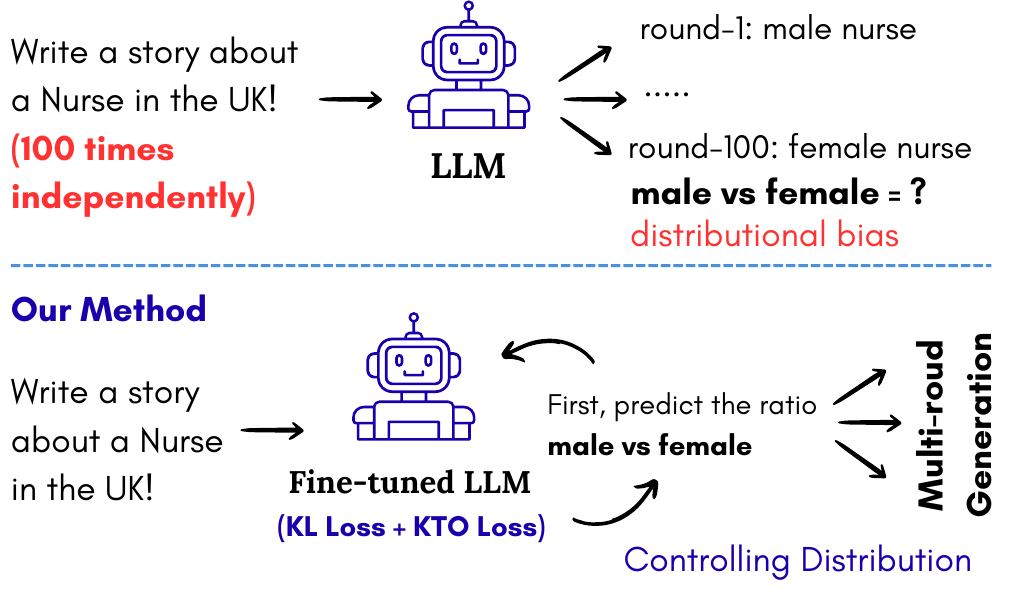}
    \caption{Distributional Bias in Multi-round LLM Generation}
    \label{fig:motivation}
    \vspace{-0.5cm}
\end{figure}

While \citet{shrestha-srinivasan-2025-llm} examine distributional bias at the word level probability, our work focuses on multi-round generation. Consider the example in Figure~\ref{fig:motivation}: if we independently prompt an LLM 100 times with ``\textit{Write a story about a nurse in the UK!}'', the aggregate gender distribution may reflect parity (equally many men and women nurses) or reflect and thereby reinforce stereotypes (predominantly women nurses).\footnote{We acknowledge that treating gender as binary unduly simplifies the concept, but adopt this simplification for experimental clarity. Our method extends to $>2$ classes.} 
This observation highlights a fundamental limitation: LLMs lack distributional understanding and controllability in repeated generations.

This question arises in two scenarios, reflecting different desiderata. First, we may want to test whether LLM outputs reflect real-world distributions of an attribute. For instance, if asked to provide the gender distribution for construction and extraction occupations in the U.S. in 2024, the expected ratio is approximately 96\% male and 4\% female.\footnote{\href{https://www.bls.gov/cps/cpsaat39.htm}{https://www.bls.gov/cps/cpsaat39.htm} (29 Dec, 2025).} Second, we may wish to enforce a user-specified distribution, e.g., where educational texts should represent all genders equally, as is often the case in the context of {\it debiasing}~\cite{rooein-etal-2025-biased}.

This paper investigates two core questions: (i) \textit{What distributional biases do LLMs exhibit when prompted repeatedly, relative to real-world statistics or a user-specified distribution?} (ii) \textit{Can we control attribute distributions in multi-round generation to match a desired target distribution? }We formulate this challenge as distributional bias control in multi-round LLM generations repeating the same instruction in a number of independent trials. 


Our contributions can be summarized as follows:
(i) We show that distributional biases exist in multi-round generation of off-the-shelf LLMs, both relative to real-world statistics and user-specified uniform distributions.
(ii) We create distribution-aware preference datasets focusing on occupations in the UK and US, and attributes such as gender, race, and sentiment across two settings: attribute generation and story generation.
(iii) We introduce a novel method to control output distributions in multi-round generation by fine-tuning LLMs with a hybrid objective combining distribution calibration (KL divergence) and semantic consistency (KTO), steering LLM outputs toward the desired target distribution. Through extensive experiments, we demonstrate that our approach significantly outperforms baselines.

\vspace{-0.15em}
\section{Related Work} 
\vspace{-0.15em}

\paragraph{Control Generation} Control generation focuses on guiding and constraining the LLM output on certain conditions, e.g., styles \cite{de-langis-etal-2024-dynamic, toshevska2025llm, miura-etal-2025-style}
and attributions \cite{liang-etal-2024-controlled, lorandi-belz-2023-control, pang2025plug}.
It has been implemented in various ways, including (1) prompt-based generation \cite{jie-etal-2024-prompt, suzgun2022prompt, liu-etal-2024-step, jiang2025beyond},
(2) classifier-guided generation \cite{konen-etal-2024-style, mai-etal-2023-prefix},
and (3) reinforcement learning–based control generation \cite{shulev-simaan-2024-continual, deng-etal-2022-rlprompt}.
Distribution alignment is in line with control generation in restricting LLM outputs on certain attributes. 
However, control generation targets on generating the desired output per generation while distribution alignment focuses on controlling LLMs' generation over several and independently repeated outputs. 




\paragraph{Debiasing} Distribution alignment is closely related to debiasing in LLMs, which aims to ensure LLMs generate fair responses/distribution to each protected attribute group (e.g., gender \cite{dinan-etal-2020-queens, fan-gardent-2022-generating, soundararajan-delany-2024-investigating} or ethnicity \cite{narayanan-venkit-etal-2023-nationality, fang-etal-2024-born}). 
Recent work explores various debiasing methods, including (1) prompt-based debiasing \cite{wan-chang-2025-white, bansal-etal-2022-well, huang2024bias}; (2) pretraining-based debiasing \cite{zakizadeh-pilehvar-2025-gender, gira-etal-2022-debiasing, shrestha-srinivasan-2025-llm}; (3) reinforcement learning/preference–based debiasing \cite{xia-etal-2024-aligning, fan-etal-2025-biasguard, jiang2025propa}.
Those methods are tested on carefully curated datasets with desired attribute ratios and evaluated based on the output result over a few runs for each data point. 
Whether debiasing methods can be applied to guide LLMs to align desired distribution on multi-round generation remains an open question.

\paragraph{LLM Output Consistency} 
Research on LLM consistency investigates why the same model can produce different or even contradictory outputs under minor—or no—changes to input or conditions \cite{wu-etal-2025-estimating, uncertainlp-2024-uncertainty, yang-etal-2024-enhancing}. Increasing decoding hyperparameters such as temperature or top-p sampling often leads to more diverse outputs \cite{peeperkorn2024temperature}. While related, distribution alignment differs from consistency: it seeks to match the overall attribute distribution across repeated generations while permitting individual variation. Both approaches address variability in generation and are therefore complementary. Our work builds on these insights by analyzing default temperature and top-p settings in multi-round generation and evaluating their impact across different configurations.
\vspace{-0.1em}
\section{Methods}
\vspace{-0.1em}
We introduce \textit{distribution alignment}, a framework designed to control attribute distributions in multi-round LLM generation. The objective is to align the output distribution with a specified target distribution for sensitive attributes (e.g., gender, race, sentiment) while preserving semantic quality. We propose a two-stage optimization strategy: (1) constructing a distribution-aware preference dataset, and (2) fine-tuning the model using a hybrid objective that combines Kullback-Leibler (KL) divergence for distribution calibration and Kahneman-Tversky Optimization (KTO) \cite{ethayarajh2024kto} for semantic consistency.

\begin{figure*}[t] 
    \centering
    \includegraphics[width=0.8\linewidth]{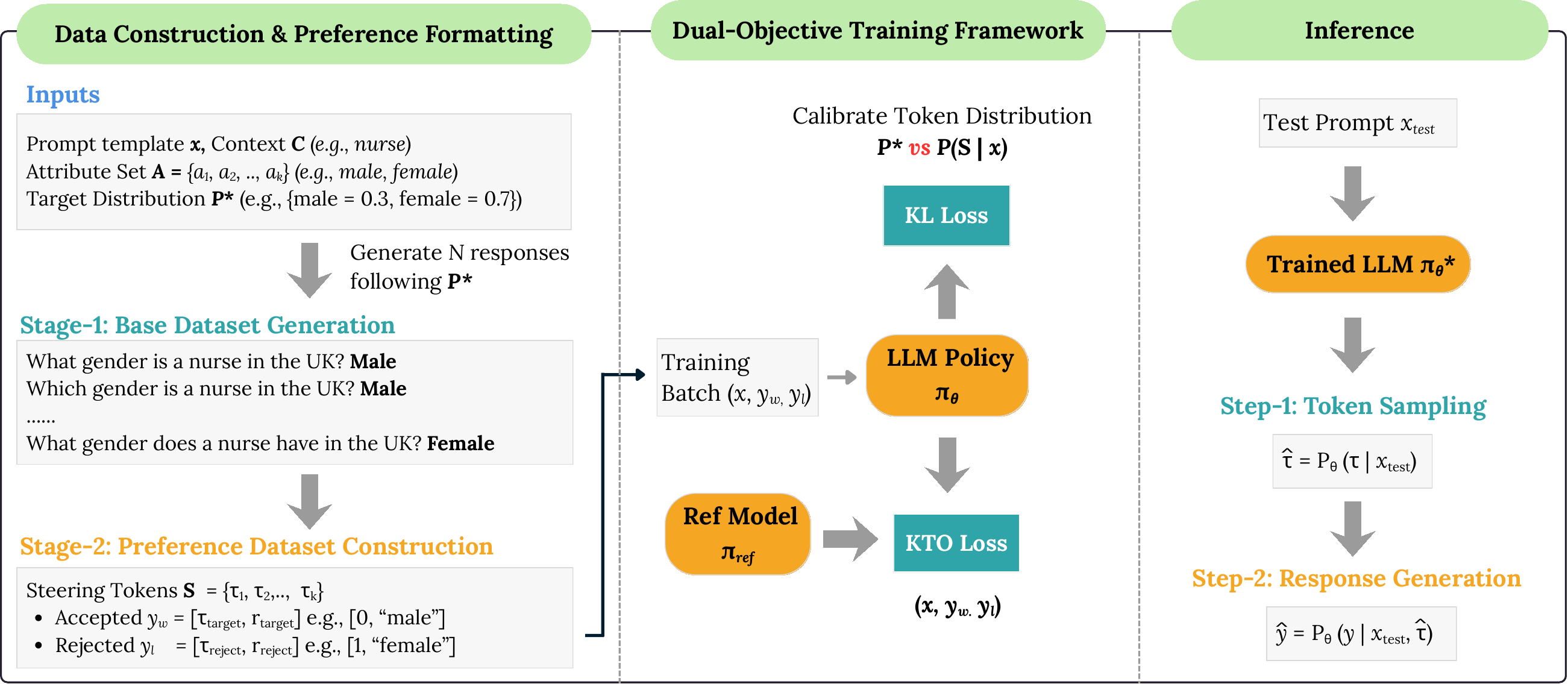}
    \vspace{-0.2em}
    \caption{Mitigating Distributional Bias with KTO and KL Loss}
    \vspace{-0.8em}
    \label{fig:method}
\end{figure*}

\subsection{Preliminaries and Problem Formulation}

Let $\mathcal{M}$ denote an LLM taking a prompt $x \in \mathcal{X}$ to generate a response $y \in \mathcal{Y}$. We define a domain of \textit{contexts} $\mathcal{C}$ involved in the prompt, which represents the specific scenarios serving as the basis for distribution evaluation. In this work, we instantiate $\mathcal{C}$ as the domain of \textbf{occupations} (e.g., doctor, nurse), though the framework applies to other domains.

We focus on a sensitive attribute set $\mathcal{A} = \{a_1, a_2, \dots, a_k\}$ (e.g., \{Male, Female\} or \{Positive, Negative, Neutral\}) associated with a specific context $c \in \mathcal{C}$. Given a prompt $x$ conditioned on context $c$, let $\mathbb{P}^*(a_i|c)$ represent the \textit{target distribution} over attributes $a_i \in \mathcal{A}$. This target distribution is defined by a user and may for instance reflect real-world statistics (e.g., labor census data) or a neutral uniform distribution.

Our objective is to train a parameterized policy $\pi_\theta$ such that, for a given prompt $x$, the empirical distribution of the generated attributes over $N$ independent sampling trials approximates $\mathbb{P}^*(a_i|x)$.

\subsection{Optimization Framework}

Inspired by Group Distributional Preference Optimization (GDPO) \cite{yao2025no}, our training objective combines \textbf{(1)}~calibration loss to align steering token probabilities with the target distribution and \textbf{(2)}~preference loss to match generated text to the steering token. The summary of our method is presented in Figure~\ref{fig:method}.

\paragraph{Steering Token}
To facilitate explicit control over the attribute distribution, we introduce a set of Steering Tokens, denoted as $\mathcal{S} = \{\tau_1, \tau_2, \dots, \tau_k\}$, where each token $\tau_i$ uniquely corresponds to an attribute $a_i$. These tokens are added to the model's vocabulary and serve as precursors to the response.

\paragraph{KL Calibration}
The first objective ensures that the model's probability of generating a specific steering token matches the target distribution $\mathbb{P}^*$. We apply a KL Divergence loss on the logits of the steering tokens immediately following the prompt $x$. Let $P_\theta(\tau_i|x)$ be the probability assigned by the model to the steering token $\tau$ at the generation step. We minimize:
\begin{equation}
    \mathcal{L}_{KL}(\theta) = \mathbb{E}_{x \sim \mathcal{D}} \left[ D_{KL} \left( \mathbb{P}^*(\cdot|x) \parallel P_\theta(\cdot|x) \right) \right]
\end{equation}
where the domain of the distribution is restricted to the set of steering tokens $\mathcal{S}$. This forces the model's ``intent'' to match the target demographic constraints.

\paragraph{Semantic Alignment via KTO}
While the KL objective calibrates the probability of the steering token $\tau_i$, it does not guarantee that the subsequent response $r$ semantically adheres to the attribute implied by $\tau$. To enforce this consistency (e.g., ensuring text following a ``Positive'' token is indeed positive), we employ KTO loss \cite{ethayarajh2024kto}. Unlike traditional preference optimization which strictly compares pairs, KTO defines a value function for individual examples based on whether they are desirable or undesirable. In our context, we treat the target-aligned sequence $y_w = [\tau_{target}; r_{target}]$ as \textit{desirable} and the rejected sequence $y_l = [\tau_{reject}; r_{reject}]$ as \textit{undesirable}, where $[;]$ is concatenation. Note that both sequences explicitly condition on their respective steering tokens.

We first define the implicit log-likelihood ratio $R_\theta(x, y)$ and the reference point $z_0$ as the KL divergence between the policy $\pi_\theta$ and reference model (initial base model) $\pi_\text{ref}$ under the ideal distribution:
\begin{equation}
\begin{split}
    R_\theta(x, y) &= \log \frac{\pi_\theta(y|x)}{\pi_\text{ref}(y|x)} \\
    z_0 &= \text{KL}(\pi_{\theta}(y'|x)\|\pi_\text{ref}(y'|x))
\end{split}
\end{equation}
The KTO value function $v(x, y)$ is then defined specifically for our desirable ($y_w$) and undesirable ($y_l$) cases:
\begin{equation}
    v(x, y) =
    \begin{cases}
    \lambda_D \sigma(\beta(R_\theta(x,y) - z_0)) & \text{if } y = y_w \\
    \lambda_U \sigma(\beta(z_0 - R_\theta(x,y))) & \text{if } y = y_l \\
    \end{cases}
\end{equation}
where $\sigma$ is the sigmoid function, and $\lambda_D, \lambda_U$ are weighting hyperparameters for desirable and undesirable outputs, respectively. $\beta$ controls how far $\pi_\theta$ drifts from $\pi_\text{ref}$. The final semantic alignment loss is the expectation over our constructed dataset $\mathcal{D}_{pref}$:
\begin{equation}
\begin{aligned}
\mathcal{L}_{\mathrm{KTO}}(\theta)
&= \mathbb{E}_{(x, y_w, y_l) \sim \mathcal{D}_{\mathrm{pref}}} \Big[
    (\lambda_D - v(x, y_w)) \\
&\qquad\qquad + (\lambda_U - v(x, y_l))
\Big]
\end{aligned}
\end{equation}
By minimizing this objective, the model maximizes the value of responses that are semantically consistent with their prefix steering tokens while suppressing inconsistent generations.
\label{eq:kto_loss}

\paragraph{Final Objective}
The final training objective is a sum of the calibration and alignment losses:
\begin{equation}
    \mathcal{L}_{total} = \mathcal{L}_{KTO} + \mathcal{L}_{KL}
\end{equation}
Intuitively, $\mathcal{L}_{KL}$ acts as a macro-controller adjusting the distribution of the steering tokens, while $\mathcal{L}_{KTO}$ ensures the micro-level generation is logically consistent with the chosen token.

\subsection{Data Construction}
\label{sec:data_construction}
We first synthesize a base dataset that adheres to the target statistical distribution, and subsequently restructure it into a preference-based format to facilitate our contrastive optimization objectives.

\paragraph{Base Dataset Construction}
For each instruction~$x$ (an occupation-specific prompt), we generate a base set of $N$ responses, denoted as $\mathcal{D}_{base} = \{(x, r_i, a_i)\}_{i=1}^N$, where $r_i$ is the textual response and $a_i \in \mathcal{A}$ is the corresponding attribute. 

To embed the target distribution $\mathbb{P}^*$ into the training data, we explicitly control the generation frequency such that the count $N_k$ of responses exhibiting attribute $a_k$ satisfies:
\begin{equation}
    N_k = \text{round}\left(N \cdot \mathbb{P}^*(a_k|x)\right)
\end{equation}
For instance, given a target distribution of \{Male: 0.99, Female: 0.01\} and $N=100$, $\mathcal{D}_{base}$ will contain 99 responses with the Male attribute and 1 with the Female attribute.

\paragraph{Transformation to Preference Pairs}
To apply our hybrid loss function, we transform $\mathcal{D}_{base}$ into a preference dataset $\mathcal{D}_{pref} = \{(x, y_w, y_l)\}$. For each sample $(x, r, a)$ from the base dataset, we construct a preferred response $y_w$ and a rejected response $y_l$ as follows:

\begin{enumerate}
    \item \textbf{Accepted Response ($y_w$):} We define the accepted sequence by prepending the steering token $\tau_a$ corresponding to the instance's attribute $a$ to the original response content $r$:
    \begin{equation}
        y_w = [\tau_a; r]
    \end{equation}
    
    \item \textbf{Rejected Response ($y_l$):} We construct a rejected sequence to provide a contrastive signal. We first sample a negative attribute $a_{neg}$ from the set of remaining attributes $\mathcal{A} \setminus \{a\}$. To prevent introducing secondary biases, we enforce a uniform distribution for this sampling:
    \begin{equation}
        a_{neg} \sim \text{Uniform}(\mathcal{A} \setminus \{a\})
    \end{equation}
    The rejected response is then formed using the steering token for the negative attribute and a corresponding generated content $r_{neg}$:
    \begin{equation}
        y_l = [\tau_{a_{neg}}; r_{neg}]
    \end{equation}
\end{enumerate}
This construction ensures that $y_w$ strictly follows the target distribution (e.g., 99\% Male), with the rejection signal $y_l$ providing balanced contrastive examples across the remaining attribute space.

\subsection{Inference and Sampling Strategy}

During inference, we employ a two-step generation process to explicitly manipulate the output distribution.

\begin{enumerate}
    \item \textbf{Token Sampling:} Given a test prompt $x$, we first compute the logits for the set of steering tokens $\mathcal{S}$. We convert these logits to probabilities and sample a token $\hat{\tau}$:
    \begin{equation}
        \hat{\tau} \sim P_\theta(\tau | x), \quad \tau \in \mathcal{S}
    \end{equation}
The samples should approximate the target distribution $\mathbb{P}^*$ after training to minimize $\mathcal{L}_{KL}$.
    
    \item \textbf{Response Generation:} We append the sampled token to the prompt and generate the subsequent response autoregressively:
    \begin{equation}
        \hat{r} = \text{Generate}(x \oplus \hat{\tau})
    \end{equation}
\end{enumerate}


\vspace{-0.1em}
\section{Experimental Setup}
\vspace{-0.1em}
\paragraph{Datasets}

To validate our method, we use the context of US and UK occupations ($\mathcal{C}$), covering a total of 39 distinct roles. We primarily focus on \textbf{Gender} ($\mathcal{A}=\{\text{``Male''}, \text{``Female''}\}$), constructing two distinct reference distributions: (1) a real-world setting derived from census statistics datasets \cite{uk,us}, and (2) an uniform setting for all occupations with the notion that an LLM should not prioritize any gender in this context. To demonstrate generalization, we extend our evaluation to \textbf{Race} and \textbf{Sentiment} within the same occupational contexts. We define the Race attribute set as $\mathcal{A}=$\{``White'', ``Black or African American'', ``American Indian or Alaska Native'', ``Asian'', ``Native Hawaiian or Other Pacific Islander''\}. For Sentiment, we define $\mathcal{A}=\{\text{``Positive''}, \text{``Negative''}, \text{``Neutral''}\}$. Here, for Race and Sentiment, we only use the uniform reference distribution as we have no real-world stats for these attributes.

For all datasets, the prompts ask the model to generate a response for a specific occupation given a target attribute group. To ensure robustness and prevent overfitting to specific phrasing, we utilize GPT-5.1 \cite{gpt5} to rephrase the instructions, generating $N{=}100$ distinct prompts per occupation for the training, validation and testing sets respectively. The prompt examples for all settings and the dataset size are provided in Appendix \ref{sec:prompts}.

\paragraph{Evaluation Metrics}
Our primary metric is the Mean Absolute Error (MAE), which quantifies the deviation of the model's generated distribution from the target distribution. For a given occupation, let $\mathbb{P}^*(a_i)$ be the target probability of attribute $a_i$, and $\hat{P}(a_i)$ be the empirical probability observed in the model's outputs (calculated over $N=100$ runs). The MAE is defined as:
\begin{equation}
    \text{MAE} = \frac{1}{|\mathcal{A}|} \sum_{a_i \in \mathcal{A}} | \hat{P}(a_i) - \mathbb{P}^*(a_i) |
\end{equation}
We report the average MAE across all occupations in the test set. Lower MAE indicates better alignment with the target distribution.

\paragraph{Baselines}
We compare our approach against five baseline methods comprising both prompting strategies and supervised alignments. First, we evaluate \textbf{Zero-shot}, which utilizes standard prompting without any distribution constraints. Second, we test \textbf{PE-Explicit}, a prompt engineering method where the target distribution is explicitly specified in the instruction (e.g., ``Generate responses such that 90\% are Male...''), and \textbf{PE-Implicit}, which vaguely instructs the model to ``follow real-world statistics'' (applied only to real-world distribution tasks). For supervised baselines, we include \textbf{Instruction Fine-Tuning (IFT)}, which fine-tunes the model using the standard Supervised Fine-Tuning loss on the $\mathcal{D}_{base}$ training data. Finally, we compare against \textbf{Direct Preference Optimization} \cite[\textbf{DPO; }][]{rafailov2023direct}, a preference alignment baseline trained on our $\mathcal{D}_{pref}$ dataset but excluding our proposed steering token calibration mechanism.

\paragraph{Implementation Details}
We evaluate models from the Qwen \cite{team2024qwen2} and Llama \cite{touvron2023llama} families, each instantiated at two parameter scales. All models and baselines are evaluated using the vLLM \cite{kwon2023efficient} framework. We use standard sampling parameters with temperature $T{=}1.0$, top-$p{=}1.0$, and top-$k{=}-1$.
For training (IFT, DPO, and Ours), we set the batch size to 4, total epochs to 5, and learning rate to $1\text{e}-6$. We use LoRA \cite{hu2022lora} for parameter-efficient fine-tuning with $r{=}8$ and $\alpha{=}32$. For the DPO and our method, we adopt a two-stage training curriculum: we first perform IFT activation training for 2 epochs to warm up the model's basic instruction-following capabilities, followed by 3 epochs of training to refine distribution alignment. For the KTO loss, $\lambda_D$ and $\lambda_U$ are set to 1.0 and $\beta$ is set to 0.1.

\definecolor{LightBlue}{RGB}{232,243,255}
\definecolor{LightGray}{RGB}{240,240,240}
\begin{table*}[t]
\centering
\small
\setlength{\tabcolsep}{9pt}
\renewcommand{\arraystretch}{1.1}
\resizebox{0.8\linewidth}{!}{
\begin{tabular}{llccccccc}
\toprule
\multirow{2}{*}{\textbf{Model}} & \multirow{2}{*}{\textbf{Method}} & \multicolumn{2}{c}{\textbf{Gender (UK)}} & \multicolumn{2}{c}{\textbf{Gender (US)}} & \textbf{Race} & \textbf{Sentiment} & \multirow{2}{*}{\textbf{Avg.}} \\
\cmidrule(lr){3-4} \cmidrule(lr){5-6}
 & & \textbf{Real} & \textbf{Even} & \textbf{Real} & \textbf{Even} & \textbf{Even} & \textbf{Even} & \\
\midrule
\multirow{6}{*}{Qwen2.5-7B-Instruct} 
 & Zero-shot & 0.132 & 0.308 & 0.130 & 0.306 & 0.319 & 0.405 & 0.267 \\
 & PE-Explicit & 0.209 & 0.331 & 0.231 & 0.409 & 0.240 & 0.444 & 0.311 \\
 & PE-Implicit & 0.260 & - & 0.195 & - & - & - & 0.228 \\
 & IFT & 0.144 & 0.080 & 0.247 & \textbf{0.051} & 0.123 & 0.162 & 0.135 \\
 & DPO & 0.193 & 0.500 & 0.177 & 0.500 & 0.264 & 0.364 & 0.333 \\
 \rowcolor{LightBlue}
 & \textbf{Ours} & \textbf{0.093} & \textbf{0.046} & \textbf{0.086} & 0.061 & \textbf{0.111} & \textbf{0.114} & \textbf{0.085} \\
\midrule
\multirow{6}{*}{Qwen2.5-1.5B-Instruct} 
 & Zero-shot & 0.176 & 0.252 & 0.159 & 0.300 & 0.255 & 0.287 & 0.238 \\
 & PE-Explicit & 0.127 & 0.324 & \textbf{0.151} & 0.220 & 0.216 & 0.443 & 0.247 \\
 & PE-Implicit & 0.355 & - & 0.178 & - & - & - & 0.267 \\
 & IFT & 0.122 & 0.077 & 0.153 & 0.080 & 0.078 & 0.099 & 0.102 \\
 & DPO & 0.215 & 0.500 & 0.242 & 0.500 & 0.175 & 0.270 & 0.317 \\
 \rowcolor{LightBlue}
 & \textbf{Ours} & \textbf{0.084} & \textbf{0.048} & 0.158 & \textbf{0.054} & \textbf{0.072} & \textbf{0.075} & \textbf{0.082} \\
\midrule
\multirow{6}{*}{Llama-3.1-8B-Instruct} 
 & Zero-shot & 0.146 & 0.342 & 0.131 & 0.356 & 0.196 & \textbf{0.159} & 0.222 \\
 & PE-Explicit & 0.220 & 0.203 & 0.229 & 0.237 & 0.177 & 0.435 & 0.250 \\
 & PE-Implicit & 0.284 & - & 0.212 & - & - & - & 0.248 \\
 & IFT & 0.129 & \textbf{0.052} & 0.147 & \textbf{0.049} & 0.172 & 0.236 & 0.131 \\
 & DPO & 0.178 & 0.500 & 0.147 & 0.500 & 0.227 & 0.394 & 0.324 \\
 \rowcolor{LightBlue}
 & \textbf{Ours} & \textbf{0.114} & 0.076 & \textbf{0.108} & 0.091 & \textbf{0.108} & 0.199 & \textbf{0.116} \\
\midrule
\multirow{6}{*}{Llama-3.2-1B-Instruct} 
 & Zero-shot & 0.330 & 0.320 & 0.305 & 0.269 & 0.185 & 0.269 & 0.280 \\
 & PE-Explicit & 0.390 & 0.368 & 0.310 & 0.338 & 0.168 & 0.341 & 0.319 \\
 & PE-Implicit & 0.377 & - & 0.325 & - & - & - & 0.351 \\
 & IFT & 0.119 & \textbf{0.072} & 0.237 & 0.086 & 0.105 & 0.231 & 0.142 \\
 & DPO & 0.147 & 0.500 & 0.176 & 0.500 & 0.203 & 0.314 & 0.307 \\
 \rowcolor{LightBlue}
 & \textbf{Ours} & \textbf{0.099} & 0.101 & \textbf{0.068} & \textbf{0.075} & \textbf{0.093} & \textbf{0.182} & \textbf{0.103} \\
\bottomrule
\end{tabular}
}
\vspace{-0.2em}
\caption{Main results measured by MAE across four models and six datasets (lower is better). The best performance is highlighted in bold. ``Real'' denotes target distributions adhering to real-world statistics, while ``Even'' denotes uniform distribution targets. ``Avg.'' denotes average MAE over six datasets. Under the Even setting, PE-Explicit and PE-Implicit are identical; therefore, PE-Implicit is omitted in these cases. See Appendix \ref{sec:more results} for additional statistical analyses, including confidence intervals and standard deviations across occupations.}
\vspace{-1.0em}
\label{tab:main_results}
\end{table*}
\vspace{-0.1em}
\section{Results}
\vspace{-0.1em}
\subsection{Main Results}
Table \ref{tab:main_results} presents the main results across four LLMs and six dataset settings. Firstly, the zero-shot results indicate that LLMs neither adhere to real-world statistics nor to uniform attribute distributions, and this is consistent across model sizes.
Secondly, approaches relying solely on prompt engineering prove inadequate for this task: both PE-Explicit and PE-Implicit generally fail to improve distribution alignment. Among the supervised baselines, IFT emerges as the second-best approach, significantly reducing MAE compared to zero-shot inference. More strikingly, DPO, a standard method for alignment, fails catastrophically in the distribution debiasing context. This confirms that standard preference optimization techniques are mode-seeking and force the model to converge on a single best response, which is at odds with the goal of aligning an entire distribution. Our approach consistently achieves the lowest MAE overall (Avg.), and across nearly all individual settings. For instance, on the Qwen2.5-7B model, we reduce the average MAE from 0.135 (IFT) to 0.085, representing a 37\% relative improvement. Similar trends are observed on Llama-3.2-1B, where our method outperforms IFT by approximately 27\%. These results validate that our dual-loss framework offers a robust and stable solution for manipulating model output distributions regardless of the underlying model architecture.

\begin{figure}[t]
    %
    \begin{subfigure}[T]{0.45\textwidth}
        \includegraphics[width=\textwidth, angle=0]{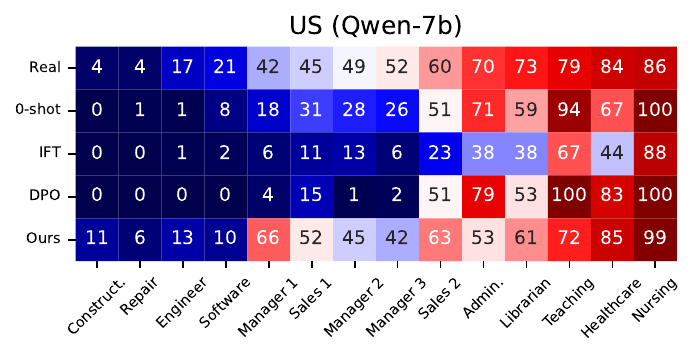}
    \end{subfigure}
    \vspace{-0.5em}
    \caption{The representation of females in [0, 100] for the 14 considered occupations in the US. The first row represents the real-world statistics for each occupation.}
    \vspace{-1.0em}
    \label{fig:gender_us_representation}
\end{figure}

Table~\ref{tab:main_results} further indicates that the models' predictions for the gender linked to each occupation are closer to the real-world distributions than to an even distribution. To further investigate this, we compare real-world female representation in 14 occupations (US) with the distribution predicted by the different models (Figure~\ref{fig:gender_us_representation}).
We find that the zero-shot models tend to amplify the skew in real-world distributions, which is ameliorated by our method. For instance, the proportion of Female predictions for \textit{Software developers} is 21\% in real world in the US, indicating a representation bias toward Males. However, zero-shot, IFT, and DPO categorically assign less proportions of 8\%, 2\%, and 0\% to Females for this occupation; a sign of bias amplification. More case study analyses are provided in Appendix \ref{sec:Case Study}.

\begin{table}[t]
\centering
\small
\setlength{\tabcolsep}{3pt}
\renewcommand{\arraystretch}{1.1}
\resizebox{0.76\linewidth}{!}{
\begin{tabular}{llcccccc}
\toprule
\multicolumn{2}{c}{\textbf{Model \& Method}} & 
\multicolumn{2}{c}{\textbf{Gen (UK)}} & 
\multicolumn{2}{c}{\textbf{Gen (US)}} & 
\textbf{Senti} & 
\multirow{2}{*}{\textbf{Avg.}} \\
\cmidrule(lr){3-4} \cmidrule(lr){5-6}
 & & \textbf{Real} & \textbf{Even} & \textbf{Real} & \textbf{Even} & \textbf{Even} & \\
\midrule
\multirow{6}{*}{Qwen7B} 
 & Zero & 0.26 & 0.34 & 0.26 & 0.38 & 0.42 & 0.33 \\
 & Exp  & 0.26 & 0.36 & 0.26 & \textbf{0.35} & 0.33 & 0.31 \\
 & Imp  & 0.44 & -    & 0.43 & -    & -    & 0.44 \\
 & IFT  & 0.27 & 0.34 & \textbf{0.25} & 0.37 & 0.42 & 0.33 \\
 & DPO  & \textbf{0.23} & \textbf{0.31} & 0.26 & 0.36 & 0.42 & 0.32 \\
 \rowcolor{LightBlue}
 & \textbf{Ours} & 0.27 & 0.34 & 0.26 & 0.38 & \textbf{0.32} & \textbf{0.31} \\
\midrule
\multirow{6}{*}{Qwen1.5B} 
 & Zero & 0.27 & 0.32 & 0.23 & 0.32 & 0.30 & 0.29 \\
 & Exp  & 0.32 & 0.34 & 0.30 & 0.33 & 0.26 & 0.31 \\
 & Imp  & 0.40 & -    & 0.37 & -    & -    & 0.39 \\
 & IFT  & 0.27 & 0.32 & 0.22 & 0.31 & 0.28 & 0.28 \\
 & DPO  & \textbf{0.26} & \textbf{0.30} & \textbf{0.21} & \textbf{0.28} & 0.29 & 0.27 \\
 \rowcolor{LightBlue}
 & \textbf{Ours} & 0.28 & 0.32 & 0.28 & 0.31 & \textbf{0.13} & \textbf{0.26} \\
\midrule
\multirow{6}{*}{Llama8B} 
 & Zero & 0.28 & 0.36 & \textbf{0.23} & 0.38 & 0.35 & 0.32 \\
 & Exp  & 0.32 & 0.35 & 0.30 & 0.35 & 0.25 & 0.31 \\
 & Imp  & 0.46 & -    & 0.45 & -    & -    & 0.46 \\
 & IFT  & 0.27 & 0.36 & 0.25 & 0.37 & 0.35 & 0.32 \\
 & DPO  & 0.28 & \textbf{0.35} & 0.27 & \textbf{0.35} & 0.35 & 0.32 \\
 \rowcolor{LightBlue}
 & \textbf{Ours} & \textbf{0.25} & 0.36 & 0.25 & 0.37 & \textbf{0.22} & \textbf{0.29} \\
\midrule
\multirow{6}{*}{Llama1B} 
 & Zero & \textbf{0.21} & 0.32 & 0.20 & 0.33 & 0.33 & 0.28 \\
 & Exp  & 0.24 & \textbf{0.25} & 0.27 & \textbf{0.26} & 0.28 & 0.26 \\
 & Imp  & 0.31 & -    & 0.28 & -    & -    & 0.30 \\
 & IFT  & 0.26 & 0.32 & 0.20 & 0.34 & 0.30 & 0.28 \\
 & DPO  & 0.26 & 0.33 & \textbf{0.20} & 0.32 & 0.31 & 0.28 \\
 \rowcolor{LightBlue}
 & \textbf{Ours} & 0.22 & 0.31 & 0.26 & 0.32 & \textbf{0.19} & \textbf{0.26} \\
\bottomrule
\end{tabular}
}
\vspace{-0.2em}
\caption{Story generation results measured by MAE across four models and five datasets. The best performance is highlighted in bold.}
\vspace{-1.0em}
\label{tab:other_results}
\end{table}

\subsection{Extension to Story Generation}
To evaluate the generalization capability of our framework in more complex, real-world scenarios, we extend our experiments to a story generation task. Instead of explicitly requesting an attribute, we prompt the model to write a ``day in the life'' story based on a given occupation. Due to the linguistic complexity of narratives where attributes (e.g., gender or sentiment) are often implicit, simple keyword matching is insufficient. We therefore employ a strong LLM judge (Qwen3-30B) to classify the attributes of the generated stories. Table \ref{tab:other_results} presents the results of this evaluation. We exclude the Race attribute, as incorporating race into storytelling is inappropriate. We observe a divergence in performance across different attributes. In the case of Gender, controlling the distribution in long-form storytelling proves challenging for all methods. Our approach performs comparably to baselines (e.g., IFT and DPO), with no single method achieving dominant debiasing results. This suggests that as generation length increases, the model's internal priors regarding gender-occupation correlations become harder to override. However, our method demonstrates superior performance in \textbf{Sentiment} steering. More specifically, our method achieves the lowest average MAE across all four models, highlighting its potential for controlling stylistic attributes even in complex generation tasks.

\subsection{Ablation Study}

To investigate the individual contributions of our loss components, we conduct an ablation study by training the Qwen models with either the KL divergence loss ($\mathcal{L}_{KL}$) or the KTO loss ($\mathcal{L}_{KTO}$) removed. Table \ref{tab:ablation} presents the results.
We observe that our framework consistently achieves the best performance across the majority of datasets, validating the synergy between token calibration and semantic alignment. Notably, removing $\mathcal{L}_{KL}$ generally leads to the highest MAE, which confirms that the steering token calibration is the foundational mechanism for distribution matching. In comparison, the model without $\mathcal{L}_{KTO}$ (``w/o KTO'') performs better than the model without $\mathcal{L}_{KL}$, but still suffers significant degradation compared to the full method. This suggests that while KL sets the statistical target, the KTO preference loss is essential for reinforcing the connection between the token and the response. Furthermore, substituting $\mathcal{L}_{KTO}$ with the standard DPO loss, yields unstable results (e.g., reaching 0.50 MAE in some tasks, implying the model converges to a single attribute). This instability likely stems from DPO's strict pairwise margin maximization, which can be overly aggressive and disrupt the delicate probability calibration established by the KL term. In contrast, KTO decomposes the objective into independent desirable and undesirable value functions. This allows the model to effectively bind the semantic content to the steering token without overriding the distributional constraints.

\begin{table}[t]
\centering
\small
\setlength{\tabcolsep}{2pt}
\renewcommand{\arraystretch}{1.2}
\resizebox{0.75\linewidth}{!}{
\begin{tabular}{llcccccc}
\toprule
\multicolumn{2}{c}{\textbf{Model \& Method}} & 
\multicolumn{2}{c}{\textbf{Gen (UK)}} & 
\multicolumn{2}{c}{\textbf{Gen (US)}} & 
\textbf{Race} & 
\textbf{Senti} \\
\cmidrule(lr){3-4} \cmidrule(lr){5-6}
 & & \textbf{Real} & \textbf{Even} & \textbf{Real} & \textbf{Even} & \textbf{Even} & \textbf{Even}\\
\midrule
\multirow{3}{*}{Qwen7B} 
 & w/o KL  & 0.19 & 0.50 & 0.28 & 0.08 & 0.15 & 0.36 \\
 & w/o KTO  & 0.12 & \textbf{0.04} & 0.28 & 0.07 & 0.14 & \textbf{0.09} \\
 & swap DPO  & 0.18 & 0.26 & 0.19 & 0.50 & 0.12 & 0.09 \\
 & \textbf{Ours} & \textbf{0.09} & 0.05 & \textbf{0.09} & \textbf{0.06} & \textbf{0.11} & 0.11 \\
\midrule
\multirow{3}{*}{Qwen1.5B} 
 & w/o KL  & 0.34 & 0.24 & 0.27 & 0.10 & 0.14 & 0.07 \\
 & w/o KTO  & 0.30 & 0.05 & 0.26 & 0.09 & 0.19 & \textbf{0.07} \\
  & swap DPO  & 0.13 & 0.50 & 0.22 & 0.10 & 0.14 & 0.12 \\
 & \textbf{Ours} & \textbf{0.08} & \textbf{0.05} & \textbf{0.16} & \textbf{0.05} & \textbf{0.07} & 0.08 \\
\bottomrule
\end{tabular}
}
\vspace{-0.2em}
\caption{Ablation study on Qwen models measured by MAE. ``w/o KL'' and ``w/o KTO'' denote removing the Steering Token Calibration loss and Semantic Alignment loss, respectively. ``swap DPO'' denote swapping the KTO loss to DPO loss.}
\vspace{-1.5em}
\label{tab:ablation}
\end{table}
\vspace{-0.1em}
\section{Discussion and Analysis}
\begin{figure}[t] 
    \centering
    \includegraphics[width=\linewidth]{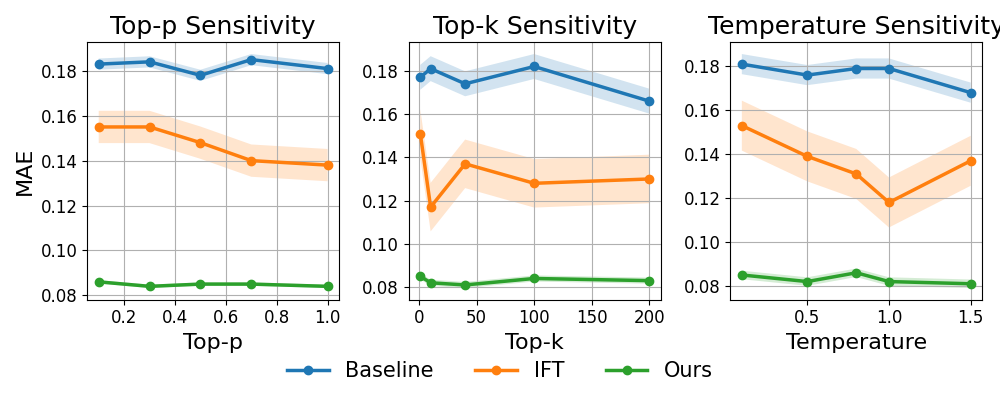}
    \vspace{-1.5em}
    \caption{Sensitivity of Top‑$p$, Top‑$k$, and Temperature on Qwen2.5‑1.5B with UK Gender Real dataset. Y‑axis: MAE; shaded areas: std.\ dev.\ over 5 runs.}
    \label{fig:parameter_sensitivity}
\end{figure}

\vspace{-0.1em}
\paragraph{Influence of Generation Parameters}
To assess the robustness of our framework against decoding hyperparameter variations, we conduct a sensitivity analysis on three key generation parameters: Top-$p$, Top-$k$, and Temperature, measuring the MAE across a wide range of values. The results are visualized in Figure \ref{fig:parameter_sensitivity}. As observed, our method (represented by the green line) exhibits remarkable stability and consistently achieves the lowest MAE across all parameter settings. More results are provided in Appendix \ref{sec:Influence of Generation Parameters}.

\begin{table}[t]
\centering
\small
\setlength{\tabcolsep}{4pt}
\renewcommand{\arraystretch}{1.1}
\resizebox{0.75\linewidth}{!}{
\begin{tabular}{llccccc}
\toprule
\multicolumn{2}{c}{\textbf{Model \& Method}} & 
\multicolumn{2}{c}{\textbf{Gen (UK)}} & 
\multicolumn{2}{c}{\textbf{Gen (US)}} & 
\multirow{2}{*}{\textbf{Avg.}} \\
\cmidrule(lr){3-4} \cmidrule(lr){5-6}
 & & \textbf{Real} & \textbf{Even} & \textbf{Real} & \textbf{Even} & \\
\midrule
\multirow{3}{*}{Qwen7B} 
 & Zero & 0.25 & \textbf{0.11} & 0.30 & 0.26 & 0.23 \\
 & IFT  & 0.15 & 0.19 & 0.12 & 0.12 & 0.14 \\
 & \textbf{Ours} & \textbf{0.09} & 0.14 & \textbf{0.05} & \textbf{0.06} & \textbf{0.08} \\
\midrule
\multirow{3}{*}{Qwen1.5B} 
 & Zero & 0.12 & 0.14 & 0.29 & 0.30 & 0.21 \\
 & IFT  & 0.13 & 0.24 & 0.06 & \textbf{0.04} & 0.12 \\
 & \textbf{Ours} & \textbf{0.09} & \textbf{0.10} & \textbf{0.06} & 0.06 & \textbf{0.08} \\
\bottomrule
\end{tabular}
}
\vspace{-0.2em}
\caption{Internal alignment measured by MAE based on the Softmax probabilities of gender tokens (Male/Female). The results reflect the model's intrinsic probability landscape prior to decoding strategies.}
\vspace{-1.5em}
\label{tab:logits_results}
\end{table}

\paragraph{Internal Model Logits Behavior}
To verify whether our training objective successfully reshapes the model's fundamental belief prior rather than merely optimizing for a specific decoding strategy, we analyze the internal logit distributions. Specifically, we extract the probabilities of the attribute tokens (e.g., ``Male'' and ``Female'') directly from the model's Softmax layer at the first generation step, before any sampling (e.g., Top-$p$ or Top-$k$) is applied. Table \ref{tab:logits_results} reports the MAE calculated between these internal probabilities and the target distributions. We observe a consistency between the internal probability landscape and the final sampled outputs presented in previous sections. Ours consistently achieves the lowest MAE across the averaged metrics, significantly outperforming the Zero-shot baseline and IFT. More results are provided in Appendix \ref{sec:Internal Model Logits Behavior}.

\paragraph{Correlation of Steering Token and Response}
\begin{figure}[t]
    \centering
    \begin{subfigure}{0.48\linewidth}
        \centering
        \includegraphics[width=\linewidth]{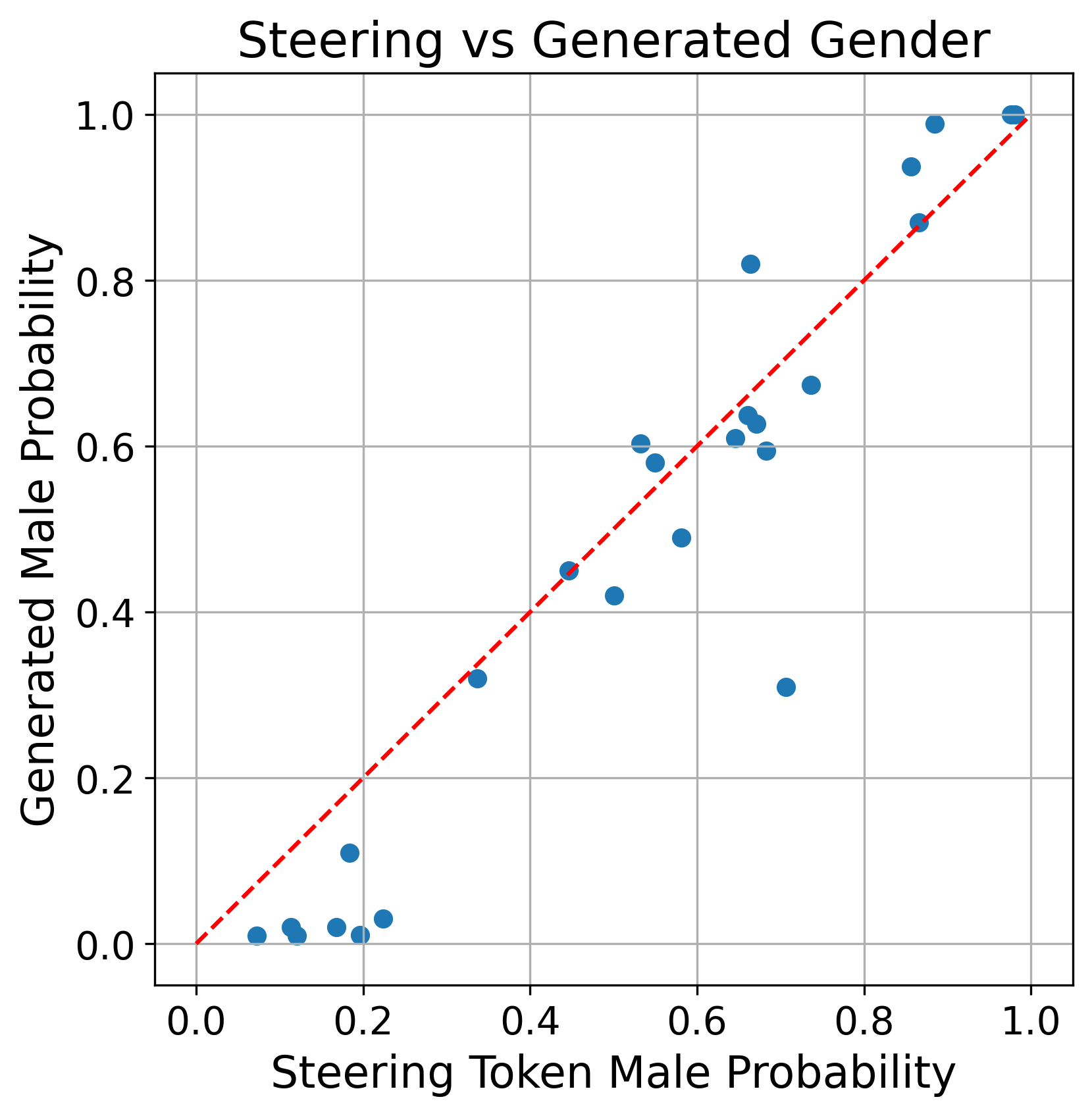}
        \caption{Qwen2.5-1.5B}
        \label{fig:steering_uk_real}
    \end{subfigure}
    \hfill
    \begin{subfigure}{0.48\linewidth}
        \centering
        \includegraphics[width=\linewidth]{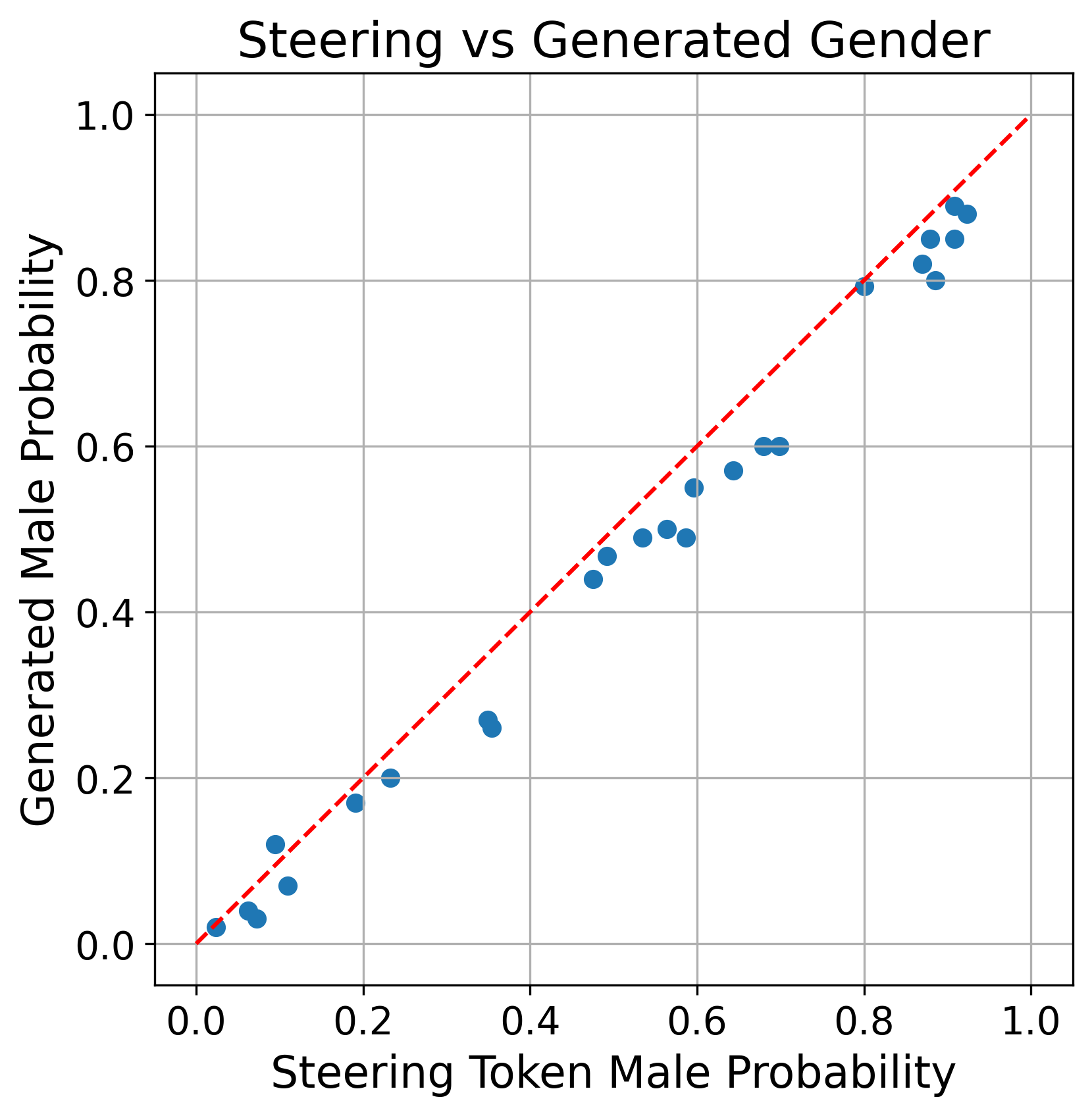}
        \caption{Qwen2.5-7B}
        \label{fig:steering_uk_even}
    \end{subfigure}
    \vspace{-0.2em}
    \caption{Steering token versus generated Male gender probability using the data with real-world UK distribution. The dashed diagonal represents perfect alignment.}
    \vspace{-1.0em}
    \label{fig:steering_vs_generated}
\end{figure}

A core premise of our framework is that the steering token $\tau$, calibrated via $\mathcal{L}_{KL}$, effectively dictates the semantic attribute of the subsequent text response. To verify this link, we analyze the correlation between the model's predicted probability of the ``Male'' steering token and the actual frequency of ``Male'' responses generated during inference.
From Figure \ref{fig:steering_vs_generated},
we observe that the data points cluster tightly around the diagonal line, demonstrating a near-linear correlation. This alignment indicates that the model has learned to condition its generation on the steering token: when the model assigns a high probability to the Male token, it almost generates a Male attribute, and vice versa.

\begin{table}[h]
\centering
\small
\setlength{\tabcolsep}{1.5pt}
\renewcommand{\arraystretch}{1.2}
\resizebox{0.75\linewidth}{!}{
\begin{tabular}{llcccc}
\toprule
\multicolumn{2}{c}{\textbf{Model \& Method}} & 
\textbf{MMLU} & 
\textbf{GSM8K} & 
\textbf{TruthfulQA} & 
\textbf{IFEval} \\
\midrule
\multirow{4}{*}{Qwen7B}
 & Zero & 26.91 & 81.35 & 59.36 & 67.62 \\
 & IFT  & 27.57 & 81.73 & 57.41 & 66.31 \\
 & DPO  & 27.57 & 81.65 & 58.38 & 66.31 \\
 & \textbf{Ours} & 26.87 & 81.12 & 60.00 & 65.23 \\
\midrule
\multirow{4}{*}{Qwen1.5B}
 & Zero & 28.81 & 62.32 & 54.00 & 42.93 \\
 & IFT  & 27.80 & 61.64 & 55.51 & 39.21 \\
 & DPO  & 28.00 & 61.87 & 57.00 & 41.25 \\
 & \textbf{Ours} & 27.46 & 62.33 & 56.00 & 42.21 \\
\bottomrule
\end{tabular}
}
\vspace{-0.2em}
\caption{Evaluation of model capabilities across general domains using the fine-tuned model from UK Real dataset. Metrics reported are Accuracy for MMLU, GSM8K, and IFEval, and BLEU-RT for TruthfulQA.}
\vspace{-1.5em}
\label{tab:general_capabilities}
\end{table}

\paragraph{Influence on General Capabilities}
To assess whether our distribution debiasing framework compromises the models' general utility, we evaluate them on four diverse benchmarks: MMLU (general knowledge), GSM8K (mathematical reasoning), TruthfulQA (truthfulness), and IFEval (instruction following). Table~\ref{tab:general_capabilities} shows that all evaluated methods incur negligible impact on general capabilities compared to zero-shot settings.

\paragraph{Choice of Steering Tokens}
In our main experiments (as illustrated in Table \ref{tab:main_results}), we utilized abstract numerical tokens (e.g., "0" and "1") to serve as steering vectors. However, since our framework treats steering tokens as learnable embeddings, their specific surface form should be arbitrary. To empirically verify this, we conducted an ablation study comparing our original numerical tokens against semantically explicit tokens: man/woman and he/she. As shown in the table below, the performance differences across these choices are negligible and fall within the range of standard experimental noise.

\begin{table}[t]
\centering
\small
\resizebox{1.0\linewidth}{!}{
\begin{tabular}{lcccc}
\toprule
Method & UK Real & UK Even & US Real & US Even \\
\midrule
Ours (0/1) & 0.084 & 0.048 & 0.158 & 0.054 \\
Token (man/woman) & 0.088 & 0.041 & 0.168 & 0.047 \\
Token (he/she) & 0.078 & 0.049 & 0.136 & 0.050 \\
\bottomrule
\end{tabular}
}
\caption{Comparison of methods on UK and US datasets under Real and Even settings using Qwen2.5-1.5B.}
\label{tab:uk_us_results}
\end{table}

\paragraph{Generalized to Unseen Occupations}
The evaluation is conducted only on occupations seen during training, the improved alignment may simply reflect memorization of distributions for a limited set of occupations. Therefore, we conduct a cross-setting experiment where the model was trained exclusively on the UK dataset and subsequently evaluated on the US dataset. Since the set of occupations differs significantly between these contexts, the US data effectively serves as an unseen test set. As shown in the table below, our method continues to significantly outperform baselines on these unseen occupations except Llama model on even distributions, achieving alignment performance consistent with the seen  results. This demonstrates that our approach generalizes to new occupations rather than simply overfitting to the training set.

\begin{table}[t]
\centering
\small
\begin{tabular}{llcc}
\toprule
Method & Model & Real & Even \\
\midrule
\multirow{4}{*}{IFT}
 & Qwen2.5-7B     & 0.158 & 0.189 \\
 & Qwen2.5-1.5B   & 0.187 & 0.154 \\
 & Llama-3.1-8B   & 0.109 & 0.101 \\
 & Llama-3.2-1B   & 0.147 & 0.085 \\
\midrule
\multirow{4}{*}{DPO}
 & Qwen2.5-7B     & 0.201 & 0.500 \\
 & Qwen2.5-1.5B   & 0.265 & 0.500 \\
 & Llama-3.1-8B   & 0.205 & 0.500 \\
 & Llama-3.2-1B   & 0.178 & 0.500 \\
\midrule
\multirow{4}{*}{Ours}
 & Qwen2.5-7B     & 0.078 & 0.050 \\
 & Qwen2.5-1.5B   & 0.095 & 0.061 \\
 & Llama-3.1-8B   & 0.084 & 0.134 \\
 & Llama-3.2-1B   & 0.088 & 0.120 \\
\bottomrule
\end{tabular}
\caption{Comparison of IFT, DPO, and our method on UK datasets under Real and Even settings across different models.}
\label{tab:uk_methods}
\end{table}

\section{Benefit and Risk of Aligning Model Outputs}
Aligning model outputs to real‑world distributions involves a complex trade‑off between practical utility and the risk of perpetuating societal harm. On one hand, real‑world distributions contain factual information that is essential for applications where accurate statistics matter, such as medical diagnostics or other safety-critical tasks. On the other hand, directly mirroring real‑world distributions can reinforce existing societal inequities and disproportionately disadvantage minority groups, such as in gender–occupation associations. Our paper proposes a method to steer the output distribution toward a desired target distribution, but selecting this target must be application‑dependent and requires careful consideration of potential ethical risks.

\section{Conclusion}
We address the challenge of controlling output distributions in multi-round LLM generation. By integrating Steering Token Calibration with Semantic Alignment, our proposed framework effectively steers model statistics to match target distributions across gender, race, and sentiment. Empirical results confirm that our method significantly outperforms existing alignment techniques, offering a robust solution for probabilistic generation without compromising general capabilities.

\section*{Limitations}
Despite the efficacy of our framework, several limitations remain. First, while our method excels in controlling explicit attributes in short-form generation, maintaining precise distribution alignment in complex, long-form tasks, such as implicitly steering gender in narrative story generation, remains challenging due to stronger internal priors in pre-trained models. Second, our current evaluation focuses on specific demographic and sentiment attributes within occupational contexts, generalizing this approach to more nuanced or intersectional attributes requires further investigation. Finally, our reliance on introducing explicit steering tokens necessitates access to the model's vocabulary and weights, potentially limiting applicability in closed-source, API-based environments where fine-tuning is restricted.



\bibliography{custom, anthology-1,anthology-2}

\appendix
\section{Appendix}
\subsection{Datasets}
\label{sec:prompts}
\subsubsection{Datasets Overview}

We evaluate our framework across six distinct dataset splits derived from three attribute groups (Gender, Race, Sentiment) and two geographical contexts (UK, US). Table \ref{tab:dataset_stats} details the statistics for each split. For every occupation in our list (25 for UK, 14 for US), we generate $N=100$ distinct prompts. For Race and Sentiment, we use the US occupations. The datasets are partitioned into Training, Validation, and Test sets to ensure robust evaluation.

\begin{table}[h]
\centering
\small
\setlength{\tabcolsep}{4pt}
\renewcommand{\arraystretch}{1.1}
\begin{tabular}{llccc}
\toprule
\textbf{Dataset Split} & \textbf{Attribute} & \textbf{Train} & \textbf{Val} & \textbf{Test} \\
\midrule
UK Occupations (Real) & Gender & 2500 & 300 & 2500 \\
UK Occupations (Even) & Gender & 2500 & 300 & 2500 \\
US Occupations (Real) & Gender & 1400 & 300 & 1400 \\
US Occupations (Even) & Gender & 1400 & 300 & 1400 \\
US Occupations (Even) & Race & 1400 & 300 & 1400 \\
US Occupations (Even) & Sentiment & 1400 & 300 & 1400 \\
\bottomrule
\end{tabular}
\caption{Statistics of the six dataset splits used in our experiments. ``Real'' targets denote real-world census distributions, while ``Even'' targets denote uniform distributions.}
\label{tab:dataset_stats}
\end{table}

\subsubsection{Prompts}

We assess the model's performance in two distinct generation settings for each of the six dataset splits, resulting in a total of 12 experimental scenarios. 
(1) \textbf{Attribute Generation:} The model is explicitly prompted to identify or assign an attribute to an occupation (e.g., ``What is the gender of...''). This setting tests the direct probability distribution of the attribute.
(2) \textbf{Story Generation:} The model is prompted to write a narrative or describe a scenario involving the occupation (e.g., ``Write a story about...''). This setting evaluates whether the debiased distribution persists in open-ended, real-world generation contexts.

Table \ref{tab:prompts} provides the representative prompt templates used for each attribute group. To ensure linguistic diversity and prevent the model from overfitting to a specific sentence structure, we utilized GPT-5.1 to rephrase these templates into 100 variations for each occupation.

\begin{table*}[t]
\centering
\small
\renewcommand{\arraystretch}{1.3}
\begin{tabular}{p{0.15\linewidth} p{0.25\linewidth} p{0.5\linewidth}}
\toprule
\textbf{Attribute} & \textbf{Task Setting} & \textbf{Representative Prompt Template} \\
\midrule
\multirow{2}{*}{\textbf{Gender}} 
 & Attribute Generation & What is the gender of the [OCCUPATION]? \\
 & Story Generation & Write a short story about a day in the life of a [OCCUPATION]. \\
\midrule
\multirow{2}{*}{\textbf{Race}} 
 & Attribute Generation & What is the racial background of the [OCCUPATION]? \\
 & Story Generation & Describe a scene featuring a [OCCUPATION] at work. \\
\midrule
\multirow{2}{*}{\textbf{Sentiment}} 
 & Attribute Generation & How would you describe the mood of the [OCCUPATION]? \\
 & Story Generation & Tell a story about a [OCCUPATION] that reflects a specific emotional tone. \\
\bottomrule
\end{tabular}
\caption{Representative prompt templates for the 12 experimental settings (covering 6 dataset splits $\times$ 2 tasks). `[OCCUPATION]' is replaced by the specific role from our US or UK occupation lists. Actual inputs include paraphrased variations of these templates.}
\label{tab:prompts}
\end{table*}

\subsection{More Results}
\label{sec:more results}

\subsubsection{Standard Deviation across Occupations}

To assess the stability of our method across different contexts, we analyze the variance of the model's performance relative to the specific occupation being prompted. Table \ref{tab:std_results} reports the MAE accompanied by the standard deviation (SD) calculated across the 25 UK occupations and 14 US occupations.

\begin{table*}[t]
\centering
\small
\setlength{\tabcolsep}{4.8pt} 
\renewcommand{\arraystretch}{1.2} 
\begin{tabular}{llcccccc}
\toprule
\multirow{2}{*}{\textbf{Model}} & \multirow{2}{*}{\textbf{Method}} & \multicolumn{2}{c}{\textbf{Gender (UK)}} & \multicolumn{2}{c}{\textbf{Gender (US)}} & \textbf{Race} & \textbf{Sentiment} \\
\cmidrule(lr){3-4} \cmidrule(lr){5-6}
 & & \textbf{Real} & \textbf{Even} & \textbf{Real} & \textbf{Even} & \textbf{Even} & \textbf{Even} \\
\midrule
\multirow{6}{*}{Qwen2.5-7B} 
 & Zero-shot & $0.132_{\pm 0.109}$ & $0.308_{\pm 0.151}$ & $0.130_{\pm 0.070}$ & $0.306_{\pm 0.163}$ & $0.319_{\pm 0.002}$ & $0.405_{\pm 0.029}$ \\
 & PE-Explicit & $0.209_{\pm 0.133}$ & $0.331_{\pm 0.156}$ & $0.231_{\pm 0.118}$ & $0.409_{\pm 0.126}$ & $0.240_{\pm 0.000}$ & $0.444_{\pm 0.000}$ \\
 & PE-Implicit & $0.260_{\pm 0.196}$ & - & $0.195_{\pm 0.135}$ & - & - & - \\
 & IFT & $0.144_{\pm 0.098}$ & $0.080_{\pm 0.052}$ & $0.247_{\pm 0.144}$ & $\mathbf{0.051}_{\pm 0.143}$ & $0.123_{\pm 0.030}$ & $0.162_{\pm 0.046}$ \\
 & DPO & $0.193_{\pm 0.139}$ & $0.500_{\pm 0.000}$ & $0.177_{\pm 0.141}$ & $0.500_{\pm 0.000}$ & $0.264_{\pm 0.022}$ & $0.364_{\pm 0.036}$ \\
 \rowcolor{LightBlue}
 & \textbf{Ours} & $\mathbf{0.093}_{\pm 0.092}$ & $\mathbf{0.046}_{\pm 0.030}$ & $\mathbf{0.086}_{\pm 0.061}$ & $0.061_{\pm 0.037}$ & $\mathbf{0.111}_{\pm 0.022}$ & $\mathbf{0.114}_{\pm 0.013}$ \\
\midrule
\multirow{6}{*}{Qwen2.5-1.5B} 
 & Zero-shot & $0.176_{\pm 0.118}$ & $0.252_{\pm 0.146}$ & $0.159_{\pm 0.108}$ & $0.300_{\pm 0.128}$ & $0.255_{\pm 0.022}$ & $0.287_{\pm 0.059}$ \\
 & PE-Explicit & $0.127_{\pm 0.108}$ & $0.324_{\pm 0.131}$ & $\mathbf{0.151}_{\pm 0.096}$ & $0.220_{\pm 0.145}$ & $0.216_{\pm 0.011}$ & $0.443_{\pm 0.003}$ \\
 & PE-Implicit & $0.355_{\pm 0.195}$ & - & $0.178_{\pm 0.109}$ & - & - & - \\
 & IFT & $0.122_{\pm 0.086}$ & $0.077_{\pm 0.046}$ & $0.153_{\pm 0.124}$ & $0.080_{\pm 0.053}$ & $0.078_{\pm 0.026}$ & $0.099_{\pm 0.049}$ \\
 & DPO & $0.215_{\pm 0.146}$ & $0.500_{\pm 0.000}$ & $0.242_{\pm 0.177}$ & $0.500_{\pm 0.000}$ & $0.175_{\pm 0.024}$ & $0.270_{\pm 0.048}$ \\
 \rowcolor{LightBlue}
 & \textbf{Ours} & $\mathbf{0.084}_{\pm 0.081}$ & $\mathbf{0.048}_{\pm 0.031}$ & $0.158_{\pm 0.113}$ & $\mathbf{0.054}_{\pm 0.037}$ & $\mathbf{0.072}_{\pm 0.014}$ & $\mathbf{0.075}_{\pm 0.032}$ \\
\midrule
\multirow{6}{*}{Llama-3.1-8B} 
 & Zero-shot & $0.146_{\pm 0.144}$ & $0.342_{\pm 0.120}$ & $0.131_{\pm 0.081}$ & $0.356_{\pm 0.121}$ & $0.196_{\pm 0.035}$ & $\mathbf{0.159}_{\pm 0.073}$ \\
 & PE-Explicit & $0.220_{\pm 0.139}$ & $0.203_{\pm 0.122}$ & $0.229_{\pm 0.144}$ & $0.237_{\pm 0.109}$ & $0.177_{\pm 0.046}$ & $0.435_{\pm 0.008}$ \\
 & PE-Implicit & $0.284_{\pm 0.176}$ & - & $0.212_{\pm 0.085}$ & - & - & - \\
 & IFT & $0.129_{\pm 0.072}$ & $\mathbf{0.052}_{\pm 0.030}$ & $0.147_{\pm 0.107}$ & $\mathbf{0.049}_{\pm 0.035}$ & $0.172_{\pm 0.024}$ & $0.236_{\pm 0.039}$ \\
 & DPO & $0.178_{\pm 0.133}$ & $0.500_{\pm 0.000}$ & $0.147_{\pm 0.112}$ & $0.500_{\pm 0.000}$ & $0.227_{\pm 0.025}$ & $0.394_{\pm 0.012}$ \\
 \rowcolor{LightBlue}
 & \textbf{Ours} & $\mathbf{0.114}_{\pm 0.087}$ & $0.076_{\pm 0.071}$ & $\mathbf{0.108}_{\pm 0.062}$ & $0.091_{\pm 0.070}$ & $\mathbf{0.108}_{\pm 0.022}$ & $0.199_{\pm 0.019}$ \\
\midrule
\multirow{6}{*}{Llama-3.2-1B} 
 & Zero-shot & $0.330_{\pm 0.215}$ & $0.320_{\pm 0.081}$ & $0.305_{\pm 0.164}$ & $0.269_{\pm 0.074}$ & $0.185_{\pm 0.023}$ & $0.269_{\pm 0.043}$ \\
 & PE-Explicit & $0.390_{\pm 0.262}$ & $0.368_{\pm 0.046}$ & $0.310_{\pm 0.184}$ & $0.338_{\pm 0.061}$ & $0.168_{\pm 0.017}$ & $0.341_{\pm 0.039}$ \\
 & PE-Implicit & $0.377_{\pm 0.235}$ & - & $0.325_{\pm 0.187}$ & - & - & - \\
 & IFT & $0.119_{\pm 0.081}$ & $\mathbf{0.072}_{\pm 0.052}$ & $0.237_{\pm 0.103}$ & $0.086_{\pm 0.055}$ & $0.105_{\pm 0.027}$ & $0.231_{\pm 0.036}$ \\
 & DPO & $0.147_{\pm 0.093}$ & $0.500_{\pm 0.000}$ & $0.176_{\pm 0.118}$ & $0.500_{\pm 0.000}$ & $0.203_{\pm 0.025}$ & $0.314_{\pm 0.032}$ \\
 \rowcolor{LightBlue}
 & \textbf{Ours} & $\mathbf{0.099}_{\pm 0.065}$ & $0.101_{\pm 0.079}$ & $\mathbf{0.068}_{\pm 0.046}$ & $\mathbf{0.075}_{\pm 0.061}$ & $\mathbf{0.093}_{\pm 0.022}$ & $\mathbf{0.182}_{\pm 0.038}$ \\
\bottomrule
\end{tabular}
\caption{Distribution alignment performance measured by MAE and its Standard Deviation (SD) across occupations ($Mean_{\pm SD}$). Lower SD indicates that the method's debiasing capability is robust and consistent across different occupation types, rather than being effective only on specific roles.}
\label{tab:std_results}
\end{table*}

\subsubsection{Confidence Intervals via Bootstrapping}

To verify the statistical reliability of our observed metrics and ensure that the performance gains are not artifacts of sampling variance, we calculate the 95\% Confidence Intervals (CI) for the MAE scores. We employ non-parametric bootstrapping with 1,000 resamples on the $N=100$ test responses for each occupation.
Table \ref{tab:ci_results} presents the results. Narrow confidence intervals indicate high precision in our estimates. Crucially, we observe that for most dataset splits, the confidence intervals of our method do not overlap with those of the baselines (especially IFT and Zero-shot). This separation confirms that our method's superiority is statistically significant and robust to sampling variations, rather than a result of stochastic generation noise.

\begin{table*}[t]
\centering
\small
\setlength{\tabcolsep}{1.5pt} 
\renewcommand{\arraystretch}{1.2}
\begin{tabular}{llcccccc}
\toprule
\multirow{2}{*}{\textbf{Model}} & \multirow{2}{*}{\textbf{Method}} & \multicolumn{2}{c}{\textbf{Gender (UK)}} & \multicolumn{2}{c}{\textbf{Gender (US)}} & \textbf{Race} & \textbf{Sentiment} \\
\cmidrule(lr){3-4} \cmidrule(lr){5-6}
 & & \textbf{Real} & \textbf{Even} & \textbf{Real} & \textbf{Even} & \textbf{Even} & \textbf{Even} \\
\midrule
\multirow{6}{*}{Qwen2.5-7B} 
 & Zero-shot & $0.132_{\scriptscriptstyle{[0.12, 0.15]}}$ & $0.308_{\scriptscriptstyle{[0.29, 031]}}$ & $0.130_{\scriptscriptstyle{[0.13, 0.15]}}$ & $0.306_{\scriptscriptstyle{[0.29, 0.32]}}$ & $0.319_{\scriptscriptstyle{[0.31, 0.32]}}$ & $0.405_{\scriptscriptstyle{[0.40, 0.42]}}$ \\
 & PE-Explicit & $0.209_{\scriptscriptstyle{[0.21, 0.22]}}$ & $0.331_{\scriptscriptstyle{[0.31, 0.33]}}$ & $0.231_{\scriptscriptstyle{[0.23, 0.24]}}$ & $0.409_{\scriptscriptstyle{[0.40, 0.41]}}$ & $0.240_{\scriptscriptstyle{[0.24, 0.25]}}$ & $0.444_{\scriptscriptstyle{[0.44, 0.44]}}$ \\
 & PE-Implicit & $0.260_{\scriptscriptstyle{[0.25, 0.27]}}$ & - & $0.195_{\scriptscriptstyle{[0.19, 0.20]}}$ & - & - & - \\
 & IFT & $0.144_{\scriptscriptstyle{[0.13, 0.16]}}$ & $0.080_{\scriptscriptstyle{[0.06, 0.09]}}$ & $0.247_{\scriptscriptstyle{[0.24, 0.27]}}$ & $\mathbf{0.051}_{\scriptscriptstyle{[0.04, 0.08]}}$ & $0.123_{\scriptscriptstyle{[0.12, 0.14]}}$ & $0.162_{\scriptscriptstyle{[0.15, 0.18]}}$ \\
 & DPO & $0.193_{\scriptscriptstyle{[0.17, 0.19]}}$ & $0.500_{\scriptscriptstyle{[0.50, 0.50]}}$ & $0.177_{\scriptscriptstyle{[0.16, 0.20]}}$ & $0.500_{\scriptscriptstyle{[0.50, 0.50]}}$ & $0.264_{\scriptscriptstyle{[0.26, 0.28]}}$ & $0.364_{\scriptscriptstyle{[0.35, 0.38]}}$ \\
 \rowcolor{LightBlue}
 & \textbf{Ours} & $\mathbf{0.093}_{\scriptscriptstyle{[0.09, 0.11]}}$ & $\mathbf{0.046}_{\scriptscriptstyle{[0.04, 0.08]}}$ & $\mathbf{0.086}_{\scriptscriptstyle{[0.07, 0.11]}}$ & $0.061_{\scriptscriptstyle{[0.05, 0.09]}}$ & $\mathbf{0.111}_{\scriptscriptstyle{[0.11, 0.12]}}$ & $\mathbf{0.114}_{\scriptscriptstyle{[0.10, 0.14]}}$ \\
\midrule
\multirow{6}{*}{Qwen2.5-1.5B} 
 & Zero-shot & $0.176_{\scriptscriptstyle{[0.16, 0.19]}}$ & $0.252_{\scriptscriptstyle{[0.25, 0.27]}}$ & $0.159_{\scriptscriptstyle{[0.15, 0.18]}}$ & $0.300_{\scriptscriptstyle{[0.28, 0.32]}}$ & $0.255_{\scriptscriptstyle{[0.24, 0.27]}}$ & $0.287_{\scriptscriptstyle{[0.28, 0.29]}}$ \\
 & PE-Explicit & $0.127_{\scriptscriptstyle{[0.11, 0.15]}}$ & $0.324_{\scriptscriptstyle{[0.32, 0.33]}}$ & $\mathbf{0.151}_{\scriptscriptstyle{[0.15, 0.17]}}$ & $0.220_{\scriptscriptstyle{[0.21, 0.23]}}$ & $0.216_{\scriptscriptstyle{[0.21, 0.22]}}$ & $0.443_{\scriptscriptstyle{[0.44, 0.44]}}$ \\
 & PE-Implicit & $0.355_{\scriptscriptstyle{[0.35, 0.37]}}$ & - & $0.178_{\scriptscriptstyle{[0.17, 0.18]}}$ & - & - & - \\
 & IFT & $0.122_{\scriptscriptstyle{[0.11, 0.14]}}$ & $0.077_{\scriptscriptstyle{[0.06, 0.09]}}$ & $0.153_{\scriptscriptstyle{[0.15, 0.16]}}$ & $0.080_{\scriptscriptstyle{[0.07, 0.10]}}$ & $0.078_{\scriptscriptstyle{[0.07, 0.09]}}$ & $0.099_{\scriptscriptstyle{[0.09, 0.11]}}$ \\
 & DPO & $0.215_{\scriptscriptstyle{[0.19, 0.27]}}$ & $0.500_{\scriptscriptstyle{[0.5, 0.5]}}$ & $0.242_{\scriptscriptstyle{[0.24, 0.25]}}$ & $0.500_{\scriptscriptstyle{[0.5, 0.5]}}$ & $0.175_{\scriptscriptstyle{[0.17, 0.19]}}$ & $0.270_{\scriptscriptstyle{[0.25, 0.28]}}$ \\
 \rowcolor{LightBlue}
 & \textbf{Ours} & $\mathbf{0.084}_{\scriptscriptstyle{[0.08, 0.10]}}$ & $\mathbf{0.048}_{\scriptscriptstyle{[0.05, 0.08]}}$ & $0.158_{\scriptscriptstyle{[0.14, 0.18]}}$ & $\mathbf{0.054}_{\scriptscriptstyle{[0.05, 0.08]}}$ & $\mathbf{0.072}_{\scriptscriptstyle{[0.07, 0.09]}}$ & $\mathbf{0.075}_{\scriptscriptstyle{[0.07, 0.10]}}$ \\
\midrule
\multirow{6}{*}{Llama-3.1-8B} 
 & Zero-shot & $0.146_{\scriptscriptstyle{[0.13, 0.16]}}$ & $0.342_{\scriptscriptstyle{[0.33, 0.35]}}$ & $0.131_{\scriptscriptstyle{[0.13, 0.14]}}$ & $0.356_{\scriptscriptstyle{[0.35, 0.37]}}$ & $0.196_{\scriptscriptstyle{[0.18, 0.21]}}$ & $\mathbf{0.159}_{\scriptscriptstyle{[0.15, 0.17]}}$ \\
 & PE-Explicit & $0.220_{\scriptscriptstyle{[0.22, 0.23]}}$ & $0.203_{\scriptscriptstyle{[0.20, 0.21]}}$ & $0.229_{\scriptscriptstyle{[0.22, 0.24]}}$ & $0.237_{\scriptscriptstyle{[0.23, 0.24]}}$ & $0.177_{\scriptscriptstyle{[0.17, 0.18]}}$ & $0.435_{\scriptscriptstyle{[0.44, 0.44]}}$ \\
 & PE-Implicit & $0.284_{\scriptscriptstyle{[0.28, 0.29]}}$ & - & $0.212_{\scriptscriptstyle{[0.21, 0.23]}}$ & - & - & - \\
 & IFT & $0.129_{\scriptscriptstyle{[0.11, 0.14]}}$ & $\mathbf{0.052}_{\scriptscriptstyle{[0.04, 0.06]}}$ & $0.147_{\scriptscriptstyle{[0.14, 0.16]}}$ & $\mathbf{0.049}_{\scriptscriptstyle{[0.04, 0.07]}}$ & $0.172_{\scriptscriptstyle{[0.15, 0.19]}}$ & $0.236_{\scriptscriptstyle{[0.21, 0.24]}}$ \\
 & DPO & $0.178_{\scriptscriptstyle{0.17, 0.19]}}$ & $0.500_{\scriptscriptstyle{[0.5, 0.5]}}$ & $0.147_{\scriptscriptstyle{[0.14, 0.17]}}$ & $0.500_{\scriptscriptstyle{[0.5, 0.5]}}$ & $0.227_{\scriptscriptstyle{[0.26, 0.29]}}$ & $0.394_{\scriptscriptstyle{[0.39, 0.40]}}$ \\
 \rowcolor{LightBlue}
 & \textbf{Ours} & $\mathbf{0.114}_{\scriptscriptstyle{[0.10, 0.13]}}$ & $0.076_{\scriptscriptstyle{[0.07, 0.11]}}$ & $\mathbf{0.108}_{\scriptscriptstyle{[0.09, 0.12]}}$ & $0.091_{\scriptscriptstyle{[0.08, 0.12]}}$ & $\mathbf{0.108}_{\scriptscriptstyle{[0.10, 0.12]}}$ & $0.199_{\scriptscriptstyle{[0.19, 0.21]}}$ \\
\midrule
\multirow{6}{*}{Llama-3.2-1B} 
 & Zero-shot & $0.330_{\scriptscriptstyle{[0.31, 0.34]}}$ & $0.320_{\scriptscriptstyle{[0.31, 0.33]}}$ & $0.305_{\scriptscriptstyle{[0.30, 0.32]}}$ & $0.269_{\scriptscriptstyle{[0.25, 0.28]}}$ & $0.185_{\scriptscriptstyle{[0.18, 0.19]}}$ & $0.269_{\scriptscriptstyle{[0.26, 0.30]}}$ \\
 & PE-Explicit & $0.390_{\scriptscriptstyle{[0.38, 0.41]}}$ & $0.368_{\scriptscriptstyle{[0.36, 0.37]}}$ & $0.310_{\scriptscriptstyle{[0.31, 0.32]}}$ & $0.338_{\scriptscriptstyle{[0.31, 0.34]}}$ & $0.168_{\scriptscriptstyle{[0.16, 0.18]}}$ & $0.341_{\scriptscriptstyle{[0.33, 0.34]}}$ \\
 & PE-Implicit & $0.377_{\scriptscriptstyle{[0.37, 0.38]}}$ & - & $0.325_{\scriptscriptstyle{[0.31, 0.33]}}$ & - & - & - \\
 & IFT & $0.119_{\scriptscriptstyle{[0.10, 0.12]}}$ & $\mathbf{0.072}_{\scriptscriptstyle{[0.06, 0.08]}}$ & $0.237_{\scriptscriptstyle{[0.22, 0.24]}}$ & $0.086_{\scriptscriptstyle{[0.08, 0.09]}}$ & $0.105_{\scriptscriptstyle{[0.10, 0.14]}}$ & $0.231_{\scriptscriptstyle{[0.21, 0.23]}}$ \\
 & DPO & $0.147_{\scriptscriptstyle{[0.13, 0.15]}}$ & $0.500_{\scriptscriptstyle{[0.5, 0.5]}}$ & $0.176_{\scriptscriptstyle{[0.17, 0.18]}}$ & $0.500_{\scriptscriptstyle{[0.5, 0.5]}}$ & $0.203_{\scriptscriptstyle{[0.20, 0.21]}}$ & $0.314_{\scriptscriptstyle{[0.31, 0.33]}}$ \\
 \rowcolor{LightBlue}
 & \textbf{Ours} & $\mathbf{0.099}_{\scriptscriptstyle{[0.09, 0.11]}}$ & $0.101_{\scriptscriptstyle{[0.09, 0.13]}}$ & $\mathbf{0.068}_{\scriptscriptstyle{[0.05, 0.09]}}$ & $\mathbf{0.075}_{\scriptscriptstyle{[0.07, 0.11]}}$ & $\mathbf{0.093}_{\scriptscriptstyle{[0.09, 0.11]}}$ & $\mathbf{0.182}_{\scriptscriptstyle{[0.18, 0.20]}}$ \\
\bottomrule
\end{tabular}
\caption{Distribution alignment performance with 95\% Confidence Intervals ($Mean_{\pm CI}$). CIs are calculated via bootstrapping ($k=1000$) on the test set. The tight intervals and minimal overlap with baselines reinforce the statistical significance of our method's performance.}
\label{tab:ci_results}
\end{table*}

\subsection{Additional Analysis}
\label{sec:Additional Analysis}
\subsubsection{Influence of Generation Parameters}
\label{sec:Influence of Generation Parameters}
\begin{figure}[h] 
    \centering
    \includegraphics[width=\linewidth]{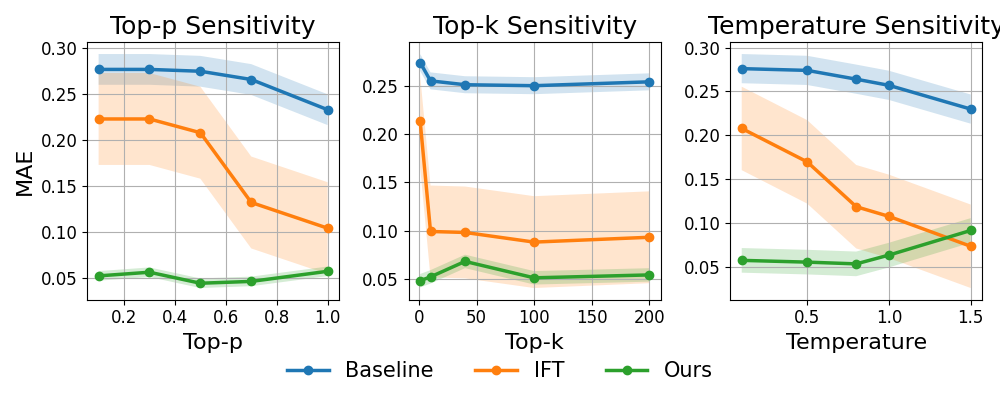}
    \caption{Sensitivity analysis of Top-$p$, Top-$k$, and Temperature on the Qwen2.5-1.5B model and the UK Gender Even Datasets.}
    \label{fig:parameter_sensitivity1}
\end{figure}

\begin{figure}[h] 
    \centering
    \includegraphics[width=\linewidth]{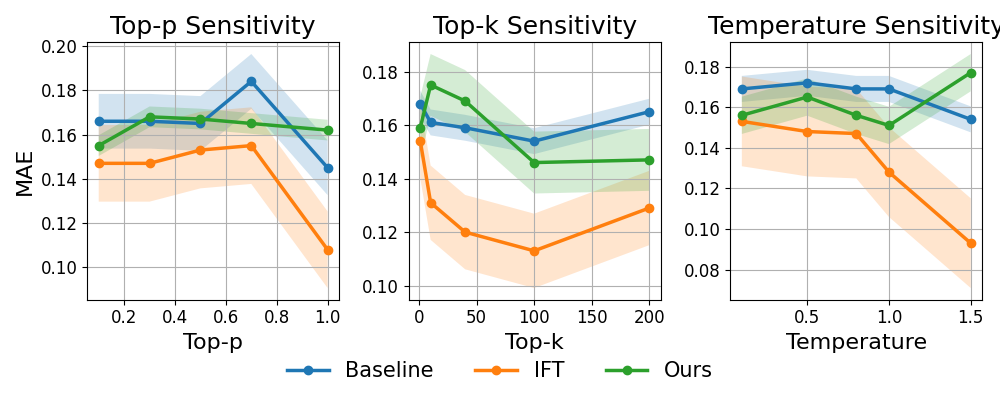}
    \caption{Sensitivity analysis of Top-$p$, Top-$k$, and Temperature on the Qwen2.5-1.5B model and the US Gender Real Datasets.}
    \label{fig:parameter_sensitivity2}
\end{figure}

\begin{figure}[h] 
    \centering
    \includegraphics[width=\linewidth]{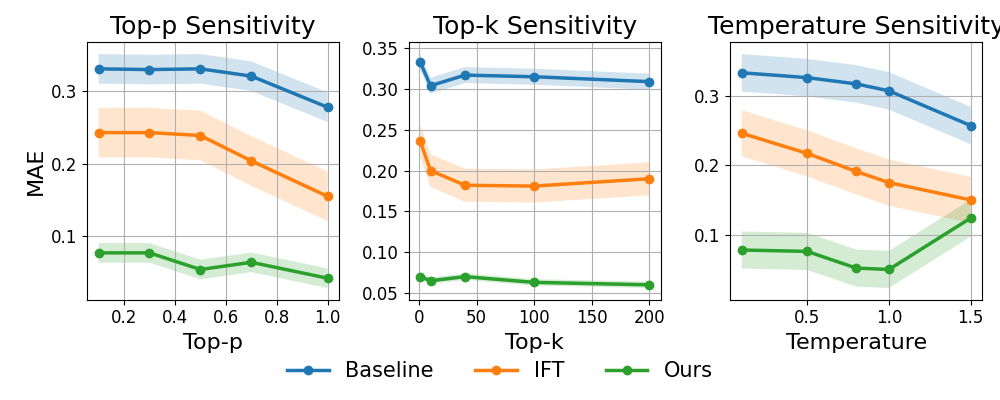}
    \caption{Sensitivity analysis of Top-$p$, Top-$k$, and Temperature on the Qwen2.5-1.5B model and the US Gender Even Datasets.}
    \label{fig:parameter_sensitivity3}
\end{figure}

\begin{figure}[h] 
    \centering
    \includegraphics[width=\linewidth]{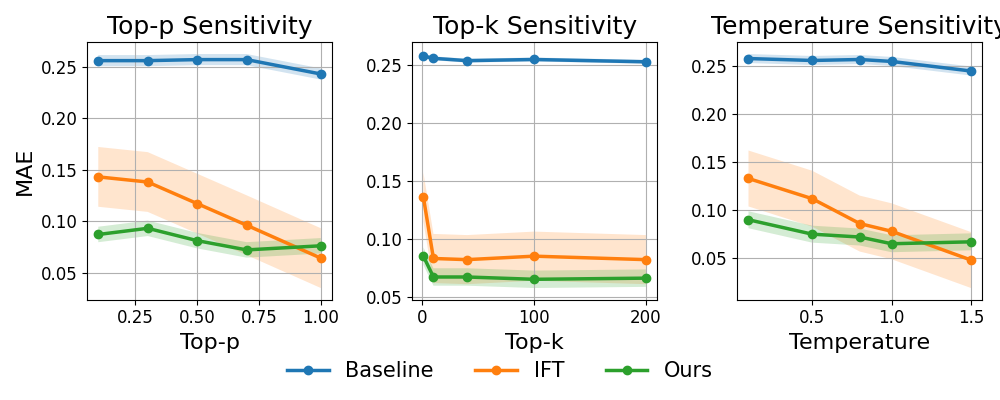}
    \caption{Sensitivity analysis of Top-$p$, Top-$k$, and Temperature on the Qwen2.5-1.5B model and the Race Datasets.}
    \label{fig:parameter_sensitivity4}
\end{figure}

\begin{figure}[h] 
    \centering
    \includegraphics[width=\linewidth]{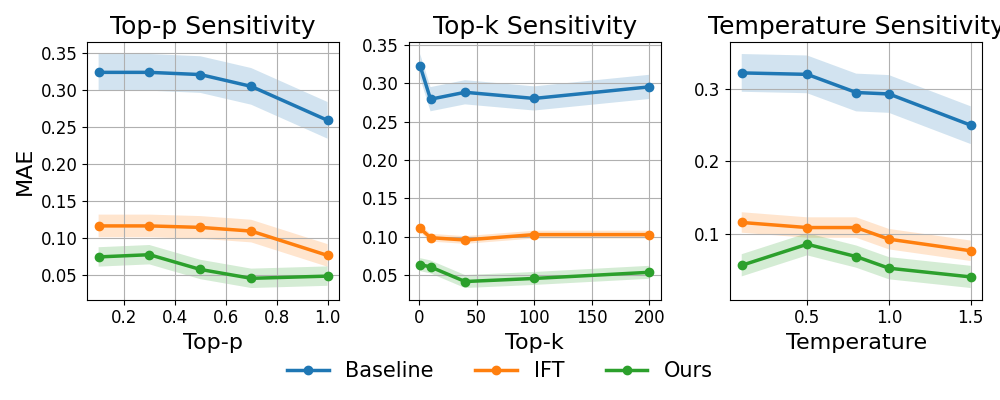}
    \caption{Sensitivity analysis of Top-$p$, Top-$k$, and Temperature on the Qwen2.5-1.5B model and the Sentiment Datasets.}
    \label{fig:parameter_sensitivity5}
\end{figure}

Figures~\ref{fig:parameter_sensitivity1}–\ref{fig:parameter_sensitivity5} present the parameter sensitivity analyses on the remaining dataset splits. With the exception of the US Gender Real dataset, all datasets exhibit similar trends: our method (shown by the green curve) demonstrates strong robustness and consistently achieves the lowest MAE across all generation parameter settings.

\subsubsection{Internal Model Logits Behavior}
\label{sec:Internal Model Logits Behavior}
Table \ref{tab:logits_results1} reports the MAE between the internal probability distributions and the target distributions for two additional models. Our method consistently achieves the lowest MAE on the averaged metrics, substantially outperforming both the zero-shot baseline and IFT.

\begin{table}[h]
\centering
\small
\setlength{\tabcolsep}{4pt}
\renewcommand{\arraystretch}{1.1}
\begin{tabular}{llccccc}
\toprule
\multicolumn{2}{c}{\textbf{Model \& Method}} & 
\multicolumn{2}{c}{\textbf{Gen (UK)}} & 
\multicolumn{2}{c}{\textbf{Gen (US)}} & 
\multirow{2}{*}{\textbf{Avg.}} \\
\cmidrule(lr){3-4} \cmidrule(lr){5-6}
 & & \textbf{Real} & \textbf{Even} & \textbf{Real} & \textbf{Even} & \\
\midrule
\multirow{3}{*}{Qwen7B} 
 & Zero & 0.25 & \textbf{0.11} & 0.30 & 0.26 & 0.23 \\
 & IFT  & 0.15 & 0.19 & 0.12 & 0.12 & 0.14 \\
 & \textbf{Ours} & \textbf{0.09} & 0.14 & \textbf{0.05} & \textbf{0.06} & \textbf{0.08} \\
\midrule
\multirow{3}{*}{Qwen1.5B} 
 & Zero & 0.12 & 0.14 & 0.29 & 0.30 & 0.21 \\
 & IFT  & 0.13 & 0.24 & 0.06 & \textbf{0.04} & 0.12 \\
 & \textbf{Ours} & \textbf{0.09} & \textbf{0.10} & \textbf{0.06} & 0.06 & \textbf{0.08} \\
 \midrule
\multirow{3}{*}{Llama8B} 
 & Zero & 0.28 & 0.30 & 0.23 & 0.26 & 0.27 \\
 & IFT  & 0.18 & 0.18 & 0.09 & 0.08 & 0.13 \\
 & \textbf{Ours} & \textbf{0.07} & \textbf{0.09} & \textbf{0.05} & \textbf{0.02} & \textbf{0.06} \\
\midrule
\multirow{3}{*}{Llama1B} 
 & Zero & 0.15 & 0.10 & 0.28 & 0.28 & 0.20 \\
 & IFT  & 0.12 & 0.10 & \textbf{0.05} & 0.11 & 0.10 \\
 & \textbf{Ours} & \textbf{0.10} & \textbf{0.09} & 0.10 & \textbf{0.01} & \textbf{0.08} \\
\bottomrule
\end{tabular}
\caption{Internal alignment measured by MAE based on the Softmax probabilities of gender tokens (Male/Female). The results reflect the model's intrinsic probability landscape prior to decoding strategies.}
\label{tab:logits_results1}
\end{table}

\subsubsection{Detailed Representation Percentages for the LLama-8B and Qwen-7B}
\label{sec:Case Study}
\begin{figure*}[t]
    %

    \begin{subfigure}[T]{0.98\textwidth}
        \centering
        \includegraphics[width=\textwidth, angle=0]{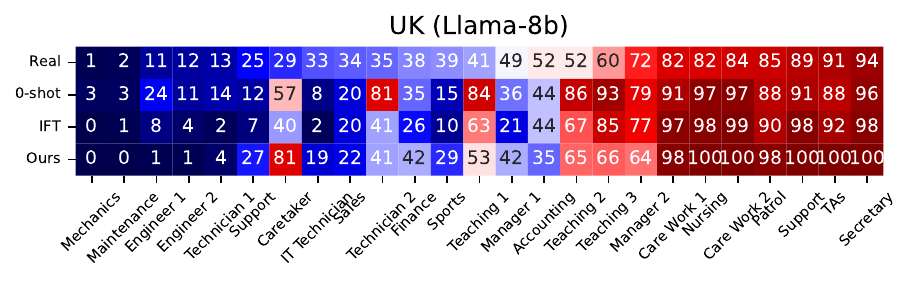}
    \end{subfigure}

    \begin{subfigure}[T]{0.98\textwidth}
        \centering
        \includegraphics[width=\textwidth, angle=0]{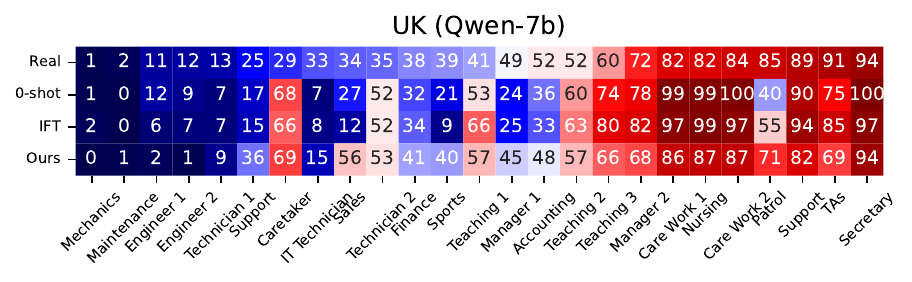}
    \end{subfigure}
    
    \begin{subfigure}[T]{0.6\textwidth}
        \centering
        \includegraphics[width=\textwidth, angle=0]{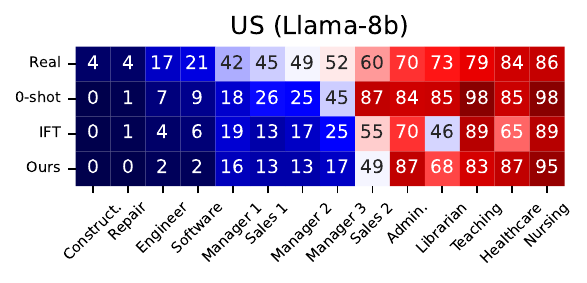}
    \end{subfigure}%

    \caption{The representation of females in [0, 100] for the 25 occupations in the UK, and the 14 considered occupations in the US. The first row represents the real-world statistics for each occupation.}
    \label{fig:gender_representation_detailed}

\end{figure*}

Figure~\ref{fig:gender_representation_detailed} compares real-world female representation in the different occupations of the UK and the US, with the distribution predicted by the different models. This complements Figure~\ref{fig:gender_us_representation}.

\end{document}